\documentclass[journal,10pt]{IEEEtran}
\usepackage{cite}
\usepackage{amsmath,amssymb,amsfonts}
\usepackage{algorithm}
\usepackage{graphicx}
\usepackage{textcomp}
\usepackage{xcolor}
\usepackage{times}
\usepackage{soul}
\usepackage{adjustbox}
\usepackage{url}
\usepackage{hyperref}
\usepackage[small]{caption}
\usepackage{graphicx}
\usepackage{amsmath}
\usepackage{amsthm}
\usepackage{booktabs}
\usepackage{amssymb}
\usepackage{amsmath}
\usepackage{subfig}
\usepackage{multirow}
\usepackage{xcolor}
\usepackage{algpseudocode}
\usepackage{color}

\DeclareMathOperator*{\argmax}{arg\,max}
\renewcommand{\algorithmicrequire}{\textbf{Input:}}  
\renewcommand{\algorithmicensure}{\textbf{Output:}} 

\def\BibTeX{{\rm B\kern-.05em{\sc i\kern-.025em b}\kern-.08em
    T\kern-.1667em\lower.7ex\hbox{E}\kern-.125emX}}

\begin{document}

\title{Adversarial Privacy Preserving Graph Embedding against Inference Attack}

\author{Kaiyang Li,
        Guangchun Luo,
        Yang Ye,
        Wei Li,~\IEEEmembership{Member,~IEEE},
        Shihao Ji,
        Zhipeng Cai,~\IEEEmembership{senior Member,~IEEE}

  \thanks{K. Li and G. Luo are with school of computer science and engneering, University of Electronic Science and Technology of China, Chengdu 600731,China (e-mail: kaiyang.li@outlook.com; gcluo@uestc.edu.cn).}
  \thanks{Y. Ye, W. Li, S. Ji, Z. Cai are  with the Department of Computer Science, Georgia State University, Atlanta, GA 30302 USA (e-mail: yye10@student.gsu.edu; wli28@gsu.edu; sji@gsu.edu; zcai@gsu.edu).}}%

\markboth{IEEE Internet of Things Journal,~Vol.~XX, No.~XX, XXX~2020}{}

\maketitle

\begin{abstract}
Recently, the surge in popularity of Internet of Things (IoT), mobile devices, social media, etc. has opened up a large source for graph data. Graph embedding has been proved extremely useful to learn low-dimensional feature representations from graph structured data. These feature representations can be used for a variety of prediction tasks from node classification to link prediction. However, existing graph embedding methods do not consider users' privacy to prevent inference attacks. That is, adversaries can infer users' sensitive information by analyzing node representations learned from graph embedding algorithms. In this paper, we propose Adversarial Privacy Graph Embedding (APGE), a graph adversarial training framework that integrates the disentangling and purging mechanisms to remove users' private information from learned node representations. The proposed method preserves the structural information and utility attributes of a graph while concealing users' private attributes from inference attacks. Extensive experiments on real-world graph datasets demonstrate the superior performance of APGE compared to the state-of-the-arts. Our source code can be found at \url{https://github.com/uJ62JHD/Privacy-Preserving-Social-Network-Embedding}.
\end{abstract}


\IEEEpeerreviewmaketitle

\section{Introduction}

Nowadays, an unprecedented volume of data generated by computer networks and services have a graph structure. For example, from network structure to control flow graph, IoT naturally generates a large amount of graph data continuously.  IoT expands human being's abilities to understand and control the physical world and grows increasingly in popularity. By the end of 2018, there were an estimated 22 billion IoT connected devices in use around the world \cite{statista.com1}.
IoT applications and services span almost all the economic and social sectors, including environment monitoring, smart industry, public safety, smart medical systems, smart grid, smart agriculture, smart transportation, behavioral patterns, etc.~\cite{gubbi2013internet,zheng2018data,zheng2020privacy}.
Another major source for graph data is social network, where people/mobile devices correspond to nodes and links represent the relationships of nodes. People use social networks to converse and connect with people sharing similar interests in the real world~\cite{leskovec2012learning,backstrom2006group,siddula2019anonymization}.
Social media has become a primary option for people to connect to new friends and interact with their current friends.
Currently, about $2.95$ billion people have social media accounts worldwide and this number will increase to almost 3.43 billion in 2023 \cite{statista.com}.

Graph embedding has made prominent progress in graph analysis in the past few years~\cite{goyal2018graph}\cite{cui2018survey}. It's a technique that learns a low-dimensional representation for each node of a graph based on network topology and nodes' attributes. There are several advantages of using graph embedding for graph analysis: (1) Through this technique, the downstream applications such as node classification~\cite{sen2008collective}, clustering~\cite{wang2017community}, link prediction~\cite{liben2007link} and graph visualization~\cite{herman2000graph} can be performed on standard machine learning algorithms whose input must be vectors. (2) Many graph mining algorithms contain iterative or combinational computations, whose complexities are typically NP-hard~\cite{gavril2011some}. By graph embedding, the computation of these tasks can be reduced dramatically. (3) It's very challenging to design parallel or distributed algorithms on graph data directly~\cite{lumsdaine2007challenges}; more efficient parallel and distributed algorithms can be developed readily on learned node representations. However, the existing graph embedding algorithms do not consider users' privacy to prevent inference attacks. That is, from the learned node representation, adversaries can infer the sensitive information that users have no intent to disclose originally \cite{ellers2019privacy}.

\begin{figure}[t]

 \centering
 \includegraphics[width=0.92\linewidth]{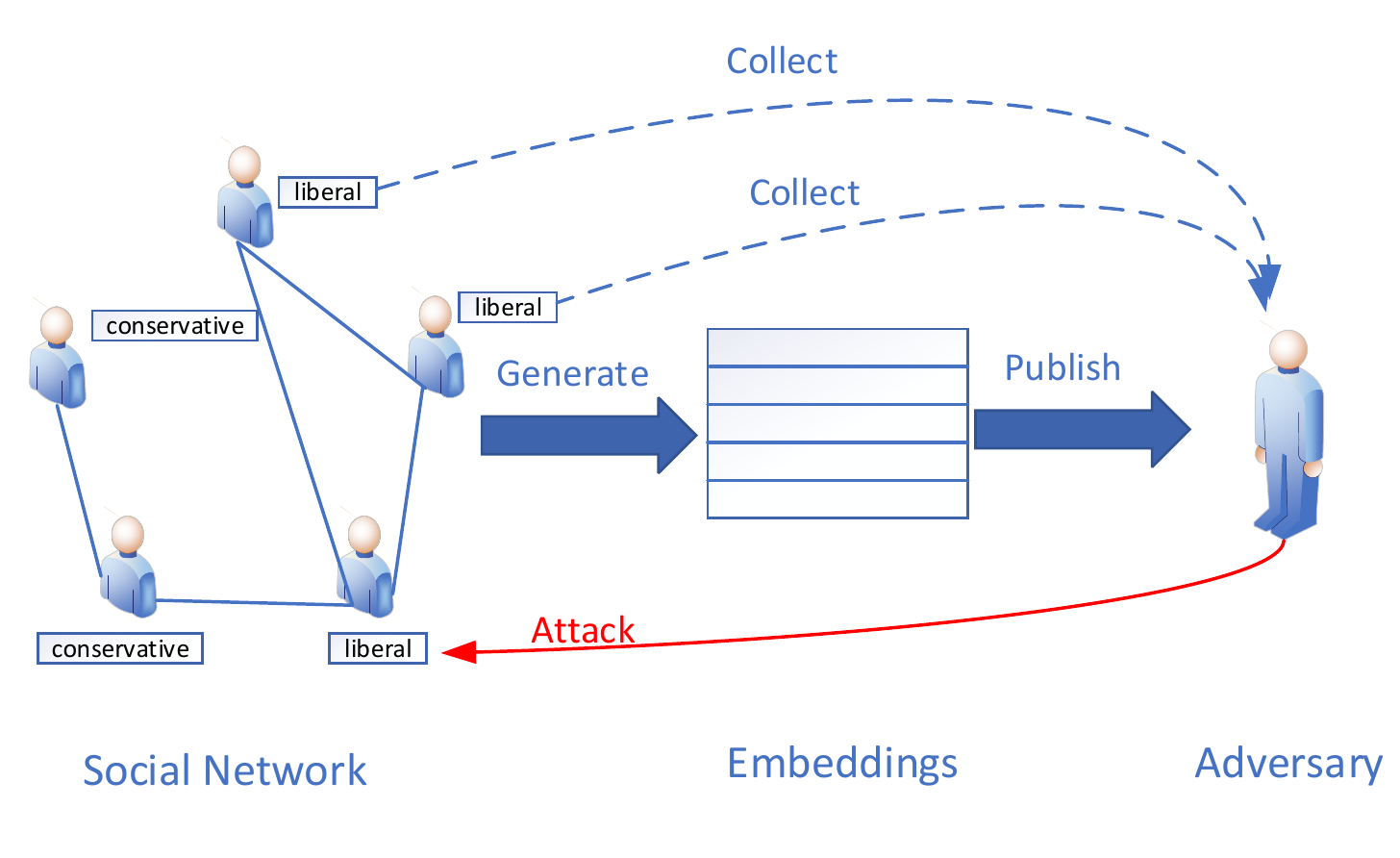}

 \caption{A example of inference attack on Graph Embedding}
 \label{Fig:attack}
\end{figure}

On the one hand,  graph data owners, such as IoT service providers and social networks, collect  a huge amount of graph data and publish data to the third parties for  research and business purposes.  The graph data can be exploited by third party analysts to predict users' purchase interest~\cite{zhang2013predicting}, discover trending events~\cite{rozenshtein2014event} and so on. However, the third parties may also use the sensitive information, such as sexual orientation and political tendency, for malicious acts. Worse, even if these sensitive attributes are deleted from graph, adversaries can still mine the privacy information via inference attacks \cite{lindamood2009inferring}\cite{heatherly2012preventing}. For examples,  Social IoT topology and users' behavior data can be used to infer users' geolocations \cite{gong2018attribute}, and users' sexual orientation could be inferred based on their Facebook data, such as users' linkages, genders, and other attributes \cite{jernigan2009gaydar}\cite{garcia2017leaking}. All of these demonstrate that our graph data is facing a serious privacy issue.
Since the vector representations of users contain  a lot of individual information, adversaries could also launch an inference attack by mining the information from the embeddings. E.g., a social network, like Facebook, publishes the embeddings of users for academic research. An adversary obtains the released embeddings and meanwhile collects part of the users' political tendency by trading~\cite{steinfeld2015trading}~\cite{acquisti2013privacy}, crawling the network~\cite{meusel2014graph}, or utilizing other resources. If the adversary trains a classifier on the embeddings and the collected political tendency, the adversary could infer the other users' political tendency via the classifier, as shown in Fig. \ref{Fig:attack}.  Therefore, before the data owners publish their graph embedding, they must preprocess their data to prevent inference attacks.

On the other hand, existing privacy protection methods typically prevent inference attacks by pruning the graph data directly \cite{gong2018attribute}\cite{cai2018collective}\cite{he2018latent}. The subsequent graph embedding on such preprocessed networks will induce suboptimal performance for privacy protection and data utility reservation. This is because:
(1) Graph embedding and privacy protection are decoupled in this process and limit the expressiveness of the model. (2) Existing methods have limited options to preprocess users' attributes and links. For example, we can either swap gender attribute of a user or delete it from the node completely to conceal users' private information. With graph embedding, users' gender information will be embedded in a continuous space and therefore we can  fine-tune the latent space through gradient updates to protect users' privacy precisely. (3) Traditional graph privacy-preserving methods only consider the first-order friendship, while graph embedding concerns global graph topology. Therefore, high-order friendship can also be used to infer users' sensitive information. For instance, two users don't have direct connections,  but they may have shared friends; if only the first-order friendship is utilized, these two users might be considered irrelevant; but they are closely related to each other due to the high-order friendship, which means their private attributes might be similar with a high probability.

What's more, the only existing graph embedding privacy preserving works ~\cite{xu2018dpne}\cite{zhang2019graph} try to achieve differential privacy on link information. Let $\mathbf{G_1}$ and  $\mathbf{G_2}$ be the neighbor graphs differing by an edge. $\mathbf{Z}_1$ and $\mathbf{Z}_2$ are the embedding matrix derived from $\mathbf{G_1}$ and $\mathbf{G_2}$, respectively. These works aim to ensure the statistical difference on $\mathbf{Z}_1$ and $\mathbf{Z}_2$ is smaller than a predefined bound.   Since the purpose of the differentially private mechanisms is not against inference attacks, these mechanisms could not defend inference attacks on sensitive attribute well.

Therefore, in this paper we propose a privacy-preserving graph embedding framework, which integrates graph embedding and privacy protection into an end-to-end pipeline against inference attack. As we will demonstrate, our method enables graph to release their privacy-preserved node embeddings to the third parties with reduced risks of privacy concern. The contributions of this paper are
\begin{itemize}
\item To the best of our knowledge, this is the first graph embedding algorithm against inference attack.
\item  To hide the sensitive information from graph embeddings, we propose two  mechanisms:  Privacy-Disentangling and Privacy-Purging from different perspectives.
\item We propose Adversarial Privacy Graph Embedding (APGE), an adversarial training based graph embedding algorithm that integrates the  mechanisms Privacy-Disentangling and Privacy-Purging to preserve users' privacy in an end-to-end graph convolution pipeline.
\item Extensive experiments on graph datasets demonstrate the superior performance of our method over previous approaches in terms of private protection and users' utility reservation.
\end{itemize}

The rest of the paper is organized as follows. Section 2 presents the problem definition and preliminary knowledge. Section 3 introduces our proposed methods.
The experimental evaluations and comparison results are reported in Section 4.
Section 5  briefly summarizes the related works on graph embedding and graph privacy preserving.
Section 6 concludes the paper.

\begin{figure*}[t]
\centering

     \subfloat[Classcial AAE\label{Classcial AAE}]{%
       \includegraphics[width=0.45\textwidth]{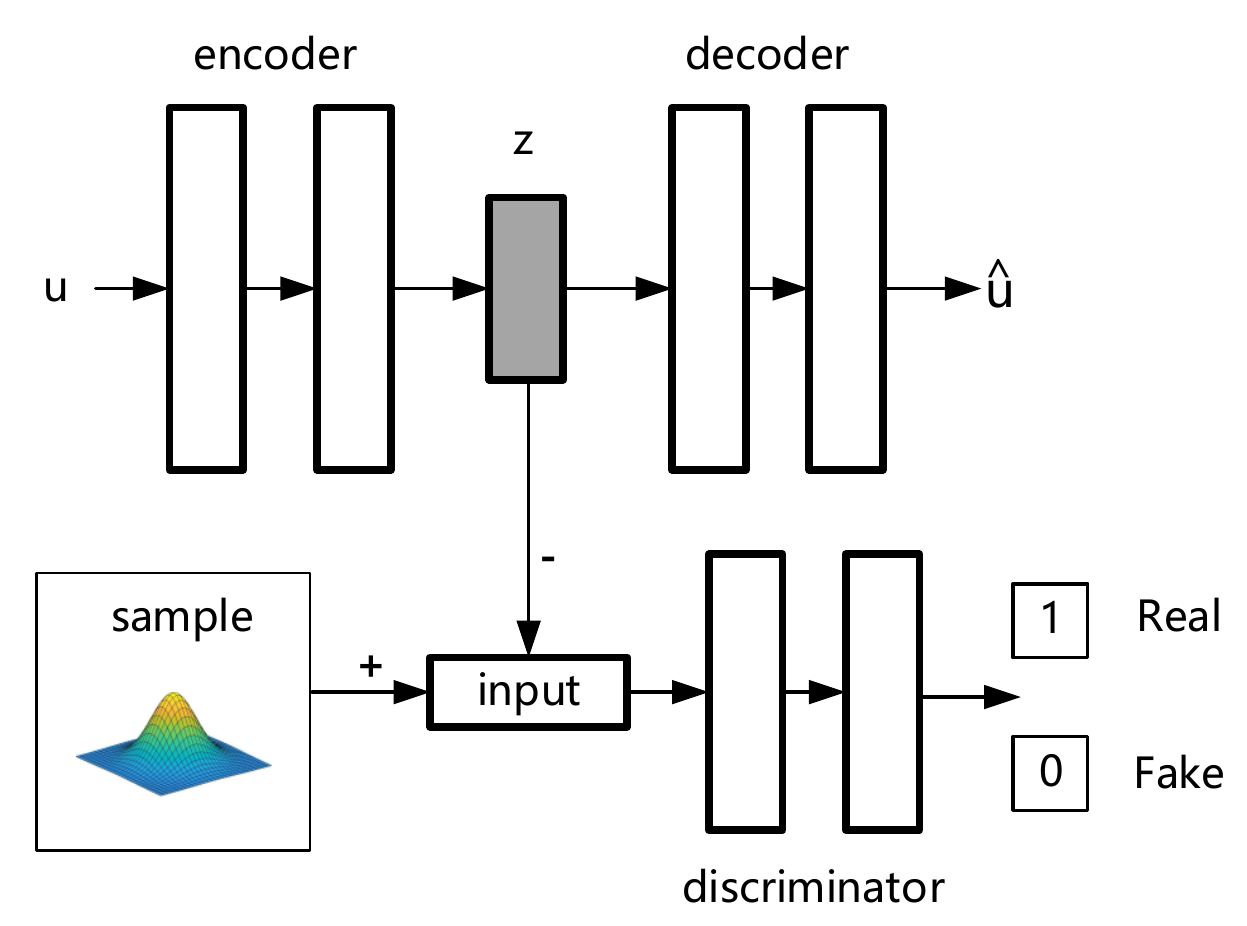}
     }
     \hspace{0.5in}
     \subfloat[Supervised AAE\label{Supervised AAE}]{%
       \includegraphics[width=0.45\textwidth]{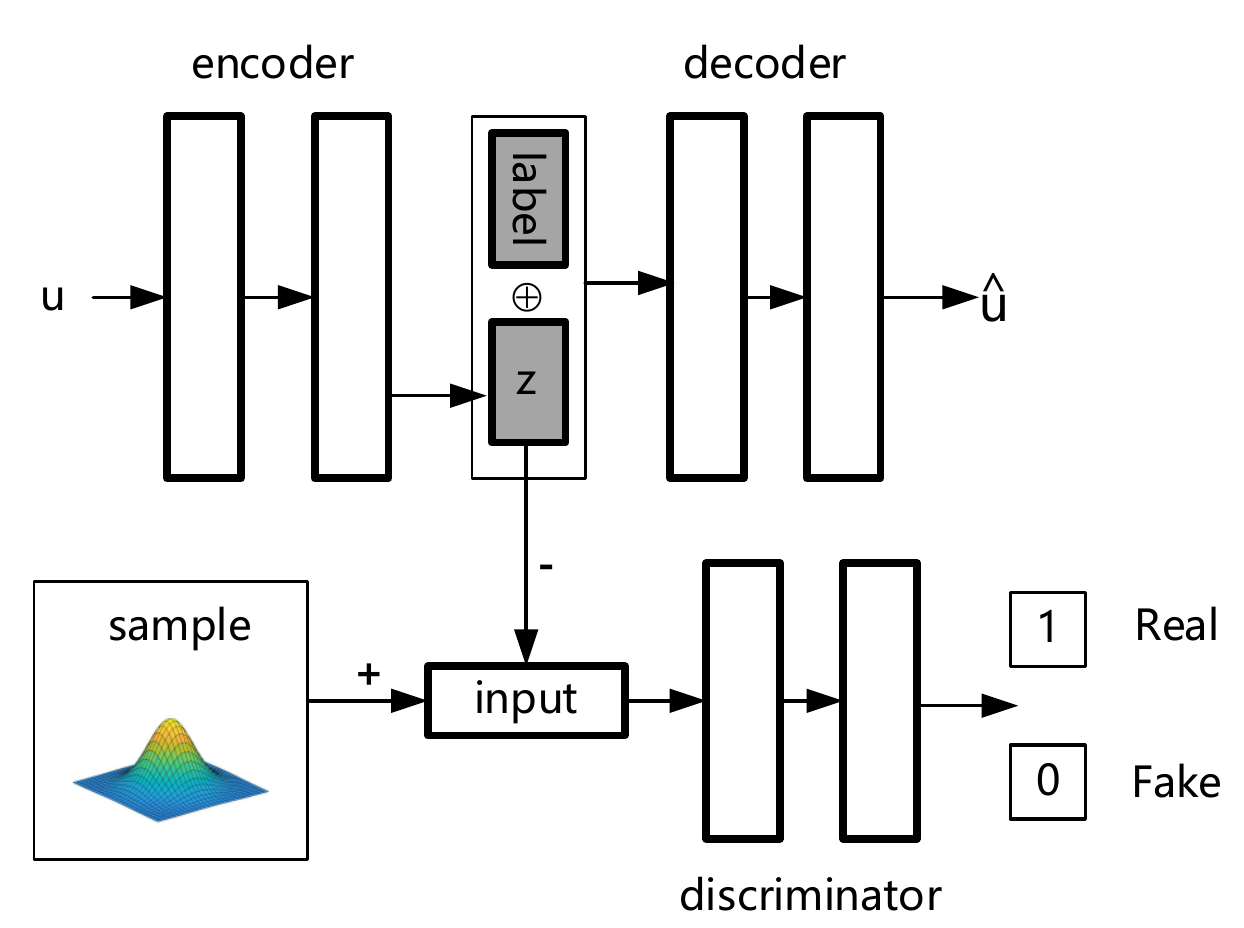}
     }
     \caption{Adversarial Autoencoder}
     \label{Fig:AAE}
\end{figure*}

\section{Problem definition and Preliminary knowledge}
In this section, we will introduce the problem definitions and preliminary knowledge of this work.
\subsection{Problem Definition}
A graph can be defined by  $\mathbf{G}(\mathbf{V},\mathbf{E},\mathbf{X})$, where $\mathbf{V}$ denotes the set of $N$ nodes, $\mathbf{E}$ denotes the set of all edges, and $\mathbf{X} \in \mathbb{R}^{N \times P}$ is the  attribute matrix, each row of which representing the feature vector of a node. Moreover, we introduce $\mathbf{A} \in \mathbb{R}^{N \times N}$ as the adjacent  matrix  of $\mathbf{G}$. In this paper, we use capital variables (e.g., $\mathbf{X}$) to denote matrices and lower-case variables (e.g., $\mathbf{x} $) to denote row vectors. Accordingly, we use $\mathbf{x}_i$ to denote the $i$-th row of the matrix $\mathbf{X}$.

\textbf{Attack model.}  We assume the adversary could obtain the released graph embeddings $\mathbf{Z}$, and the set of partial users' private attribute labels $\mathcal{Y}^{p}_{k}$. Specifically, $\mathcal{Y}^{p}_{k}$ could be captured by trading~\cite{steinfeld2015trading}, crawling~\cite{meusel2014graph}, hacking~\cite{steinmetz2015doesn}, etc. The adversary trains a classifier on $\mathbf{Z}$ and $\mathcal{Y}^{p}_{k}$ to predict the other users' private attribute as follows:
\begin{equation}
\hat{y}^{p}_i = \argmax_{y^{p}_i}P_{ \Lambda}(Y^{p}_i = y^{p}_i|\mathbf{Z},\mathcal{Y}^{p}_{k} ),
\end{equation}
 where $\Lambda$ denotes the attacking classifier, $Y^{p}_i$ is the random variable of the $i$-th user's sensitive attribute, $y^p_i$  and $\hat{y}^{(p)}_i$ denote the possible value and the predicting result of the $i$-th user's sensitive attribute, respectively. For generality, we do not constrain the types and parameters of the classifier $\Lambda$.  Note that the differential privacy is to protect the information that if a record is in the dataset, therefore the differential privacy could not defend the attack method we defined.

\textbf{Privacy and Utility.}  In this work, we define the sensitive attributes of users as privacy and the topology of the graph and users' non-sensitive attributes as utility.  That means the learned  low dimensional node representations $\mathbf{Z}$ should satisfy two properties: (1) the network structure and node utility attributes are preserved; and (2) the sensitive information is concealed, which means the privacy preserving graph embedding is robust to all kinds of inference attacks regardless the attacking classifier's types and parameters. Therefore, if we use the learned node representation matrix $\mathbf{Z}$ to analyze graph, the  accuracy of link prediction and utility attribute classification should be preserved, while the inference accuracy of private attribute should be reduced.


\subsection{Graph Autoencoder}
Graph autoencoder (GAE) \cite{kipf2016variational} embeds a graph $\mathbf{G}(\mathbf{V},\mathbf{E},\mathbf{X})$ to a low-dimensional space. The encoder of GAE is a graph convolutional network (GCN) \cite{kipf2016semi}, which updates hidden layer of a graph by
\begin{equation}
\mathbf{H}^{(l+1)} = \sigma(\mathbf{L}\mathbf{H}^{(l)}\mathbf{W}^{(l)}),
\end{equation}
where $\mathbf{L}=\mathbf{D}^{-\frac{1}{2}}\mathbf{\widetilde{A}}\mathbf{D}^{-\frac{1}{2}}$ is the symmetrically normalized graph Laplacian, $\mathbf{H}^{(l)}$ is the output of $l$-th graph convolutional layer, $\mathbf{W}^{(l)}$ is the weight matrix of $l$-th layer that is to be learned during training, and $\sigma$ represents the activation function, such as ReLU. Additionally, $\mathbf{\widetilde{A}}$ denotes the adjacency matrix $\mathbf{A}$ with diagonal elements set to 1, i.e., every node contains a self-loop, and $\mathbf{D}_{ii}=\sum_j \mathbf{\widetilde{A}}_{ij} $ denotes the degree of node $i$. Usually, we stack two graph convolutional layers as the encoder, with the propagation rule of the encoder expressed as follows:
\begin{equation}
 GCN(\mathbf{A},\mathbf{X}) = \mathbf{L}\sigma(\mathbf{L}\mathbf{X}\mathbf{W}^{(0)})\mathbf{W}^{(1)}.
\label{Eq:GCN}
\end{equation}

The purpose of GAE is to embed the structural information of a graph to a low-dimensional space. Therefore, the decoder of GAE reconstructs adjacency matrix via the inner product of embedding matrix $\mathbf{Z}$.
The reconstructed adjacency matrix $\mathbf{\hat{A}}$ and the embedding matrix $\mathbf{Z}$ can be represented as follows:
\begin{equation}
\mathbf{\hat{A}} =\sigma(\mathbf{ZZ}^T),  \quad \text{with} \quad   \mathbf{Z}=GCN(\mathbf{A},\mathbf{X}).
\label{Eq:innerP}
\end{equation}

To preserve the structural information of a graph, GAE optimizes the link prediction by minimizing the cross-entropy loss:
\begin{equation}
L_{link} = -\frac{1}{N^2}\sum_{i=1}^N \sum_{j=1}^N  \mathbf{A}_{ij} \log(\mathbf{\hat{A}}_{ij}),
\label{Eq:LossLink}
\end{equation}
where $\mathbf{A}_{ij}$ and $\mathbf{\hat{A}}_{ij}$ are the elements of  $\mathbf{A}$ and $\mathbf{\hat{A}}$, respectively.

\subsection{Adversarial Autoencoder}\label{sec:AAE}
Adversarial Autoencoder (AAE) \cite{makhzani2015adversarial} is a probabilistic autoencoder which uses the generative adversarial network (GAN) to perform variational inference by enforcing the posterior distribution of the hidden code to a specified  prior distribution. The classical AAE architecture is shown in Fig.~\ref{Fig:AAE}\subref{Classcial AAE}. The top row is an autoencoder which encodes the image $\mathbf{u}$ to hidden representation $\mathbf{z}$ and decodes $\mathbf{z}$ to a reconstructed image $\mathbf{\hat{u}}$. The bottom row is a discriminator which classifies the input is the code $\mathbf{z}$ or the noise sampled from the specified prior distribution. As we train AAE, we update the encoder and decoder to minimize the reconstruction error firstly. Then we update the discriminator to distinguish if the input is from true sample or the code generator, and we update the encoder to fool the discriminator.

In \cite{makhzani2015adversarial}, the above classical AAE has been further extended to supervised AAE that augments the decoder with the one-hot encoding of the label as shown in Fig.~\ref{Fig:AAE}\subref{Supervised AAE}. As the label and $\mathbf{z}$ are utilized jointly for reconstruction, $\mathbf{z}$ should contain the label-invariant information of input $\mathbf{u}$ for reconstruction. In other words, in supervised AAE the label information is disentangled from the latent representation $\mathbf{z}$. In this paper, such the ability of disentangled of supervised AAE is exploited for privacy protection.





\section{Proposed Model}
We first propose two privacy preserving embedding models corresponding to  mechanisms Privacy-Disentangling and Privacy-Pruning, respectively. Then we integrate the two models into one end-to-end pipeline to achieve the best performance. In the sequel, we will introduce these models one by one.

\begin{figure*}[t]
\centering

     \subfloat[Original APDGE\label{Original APDG}]{%
       \includegraphics[width=0.41\textwidth]{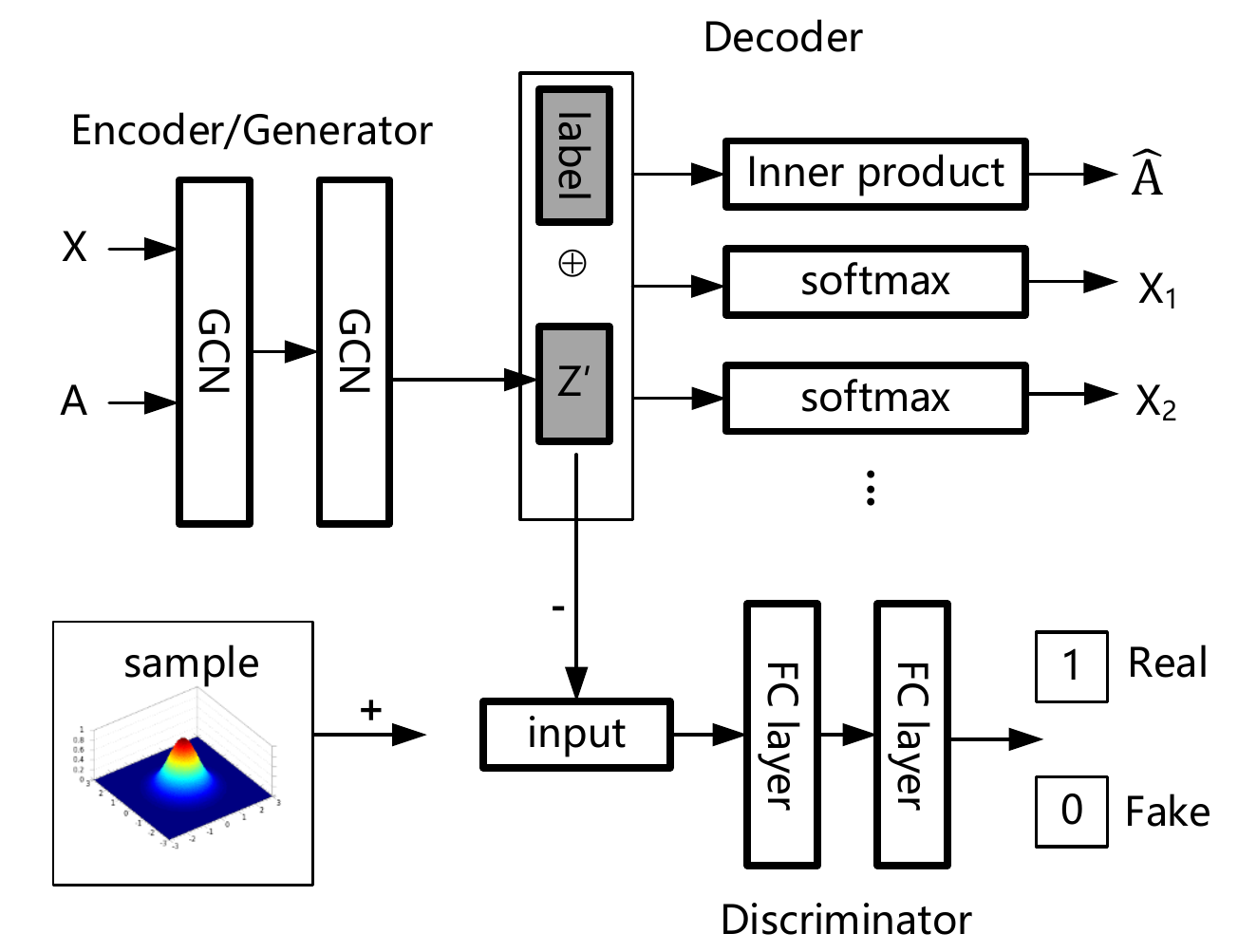}
     }
     \hspace{0.3in}
     \subfloat[Final APDGE\label{final APDGE}]{%
       \includegraphics[width=0.45\textwidth]{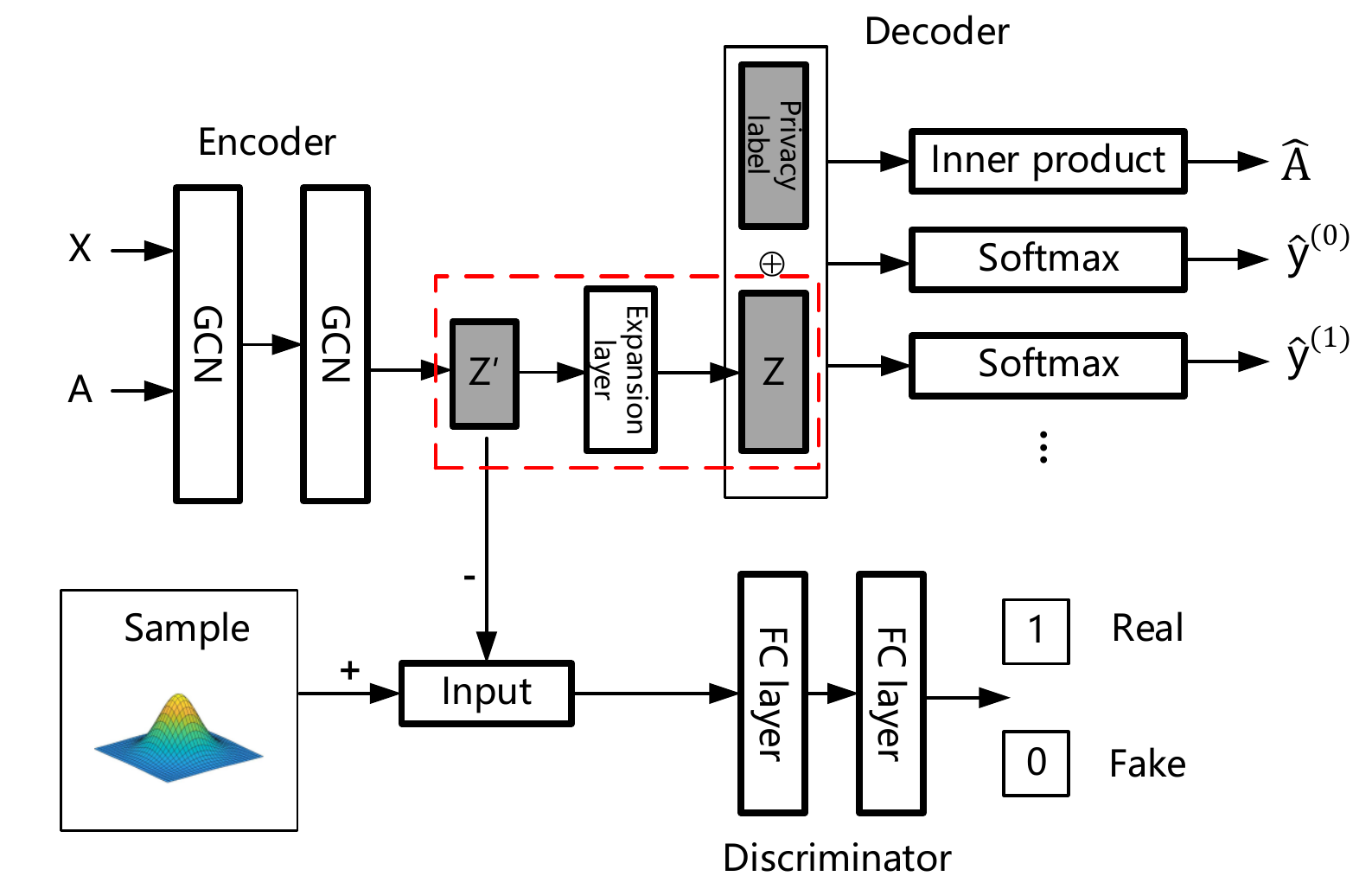}
     }
     \caption{Adversarial Privacy-Disentangled Graph Embedding}
     \label{Fig:APDGE}
\end{figure*}
\subsection{Adversarial Privacy-Disentangled Graph Embedding}\label{sec:APDGE}
As discussed in section \ref{sec:AAE}, supervised AAE can be readily adapted for privacy protection. The main issue is that the input of supervised AAE is regular grid data, such as images, while we are interested in graph structured data which are irregular. We therefore extend GAE and supervised AAE and propose  Privacy-Disentangling mechanise for privacy preserving.

The architecture of original "Adversarial Privacy-Disentangled Graph Embedding" (APDGE) is shown in Fig.~\ref{Fig:APDGE}(a), which is similar to supervised AAE and consists of encoder, decoder and discriminator. To support graph structured data, the encoder of AAE is replaced by GCN, that is, latent code $\mathbf{Z'}$ is extracted by a GCN: $\mathbf{Z'}=GCN(\mathbf{A},\mathbf{X})$ as shown in Eq.~\ref{Eq:GCN}. Note that, the encoder is the stack of two graph convolutional layers, so both first and second order friendships are considered in this model.  To disentangle privacy from latent code $\mathbf{Z'}$, we incorporate the privacy label (e.g., gender) to the decoder. On the one hand, the distribution of $\mathbf{Z'}$ is regularized to the specified  prior distribution. So the information of $\mathbf{Z'}$ is enforced to be as little as possible. On the other hand, $\mathbf{Z'}$ should contain the sufficient information to reconstruct the topology and utility attributes of the graph. In this way, the private information, which has been provided in the privacy label, is removed from $\mathbf{Z'}$. In other words, the model learns a compressed $\mathbf{Z'}$ that contains as little information of the privacy label as possible, while being sufficient for the decoder to reconstruct the utilities of graph.
Therefore, the learned graph embedding $\mathbf{Z'}$ retains the utilities of the graph while preserves the privacy information.

 We hope the dimension of $\mathbf{Z}'$ is low such that the privacy could be extruded as much as possible. However, if we directly concatenate $\mathbf{Z}'$ and the one-hot privacy label encoding as the input of decoder, the performance of decoder will be suffering because the dimension of the concatenated vector is too low. Besides, as the dimension of $\mathbf{Z}'$ could be changed to trade off the privacy and utility (which will be analyzed in section \ref{sec:hyperParameters}), we hope the dimension of the embeddings does not depend on the dimension of $\mathbf{Z}'$.  From extensive experiments,   we find a simple trick that successfully address these issues.  That is, before feeding the low dimensional representation $\mathbf{Z}'$ to the decoder, we map it to a higher dimensional $\mathbf{Z}$ via a expansion layer; then we concatenate  privacy label  with $\mathbf{Z}$ to get $\mathbf{Z^+}$, which is fed to the decoder. Specifically, we use a fully connected layer to  implement the  expansion layer.  Therefore, we modify the original APDGE to the final APDGE as shown in Fig.~\ref{Fig:APDGE}(b).

Since the learned embedding $\mathbf{Z}'$ should retain graph topology and utility attributes information, we augment the decoder to have multiple modules for adjacency matrix reconstruction and attribute classifiers. The adjacency reconstruction module calculates the inner production of $\mathbf{Z^+}$ to reconstruct adjacency matrix with the loss function for link prediction as Eq.\ref{Eq:LossLink}. For utility attribute classification, we decode $\mathbf{Z^+}$ using the softmax function, followed by the cross-entropy loss. Precisely, we predict the $i$-th user's $c$-th utility attribute by $\mathbf{\hat{y}}_i^{(c)} = \text{softmax}_c(\mathbf{z}_i^+)$, which is evaluated by the cross-entropy loss as:
   \begin{equation}
   L_{y_c} = -\frac{1}{|\mathbf{V}^{(c)}|}\sum_{i \in \mathbf{V}^{(c)}}\sum_{j=1}^{M_c}\mathbf{y}^{(c)}_{ij}\log\mathbf{\hat{y}}^{(c)}_{ij},
   \label{Eq:LossAttr}
   \end{equation}
where $\mathbf{V}^{(c)}$ is the set of users on which the $c$-th utility labels are available for training, ${\mathbf{y}^{(c)}_{ij}}$ and ${\mathbf{\hat{y}}^{(c)}_{ij}}$ are the $j$-th dimension  value of the $i$-th user's one-hot attribute label vectors  $\mathbf{y}_i^{(c)}$ and the predicting result $\mathbf{\hat{y}}_i^{(c)}$ respectively , and  $M_c$ is the dimension of the  $c$-th utility attribute. The total loss of APDGE is the combination of the link prediction loss and utility attribute classification loss:
   \begin{equation}\label{Eq:loss_rec}
   L_{recon} = L_{link} + \sum_{c \in \mathbf{C}} L_{y_c},
   \end{equation}
where $\mathbf{C}$ is the set of  utility attributes.

The discriminator $D$ of this model is the same as the discriminator of AAE. We use two fully connected layers to classify the input to be real or fake depending on if the input is a  sample $\mathbf{s}$ from a unit Gaussian distribution or $\mathbf{z'_i}$  from the encoder of APDGE. We optimize the discriminator to minimize the following loss:
   \begin{equation}
   L_{dc} = - \log(D(\mathbf{s})) - \log(1-D(\mathbf{z'_i})).
   \label{Eq:LossDc}
   \end{equation}

Overall, the training of APDGE contains three stages: (a) Update the encoder and decoder to minimize the loss Eq.~\ref{Eq:loss_rec}, (b) Update the discriminator to distinguish the real samples from the fake samples by minimizing the loss Eq.~\ref{Eq:LossDc}, and (c) Update the encoder to fool the discriminator by maximizing the loss Eq.~\ref{Eq:LossDc}. Upon finishing the training, we get the privacy-preserved node representation (latent code) matrix $\mathbf{Z}$.

\begin{figure}[htb]

 \centering
 \includegraphics[width=0.75\linewidth]{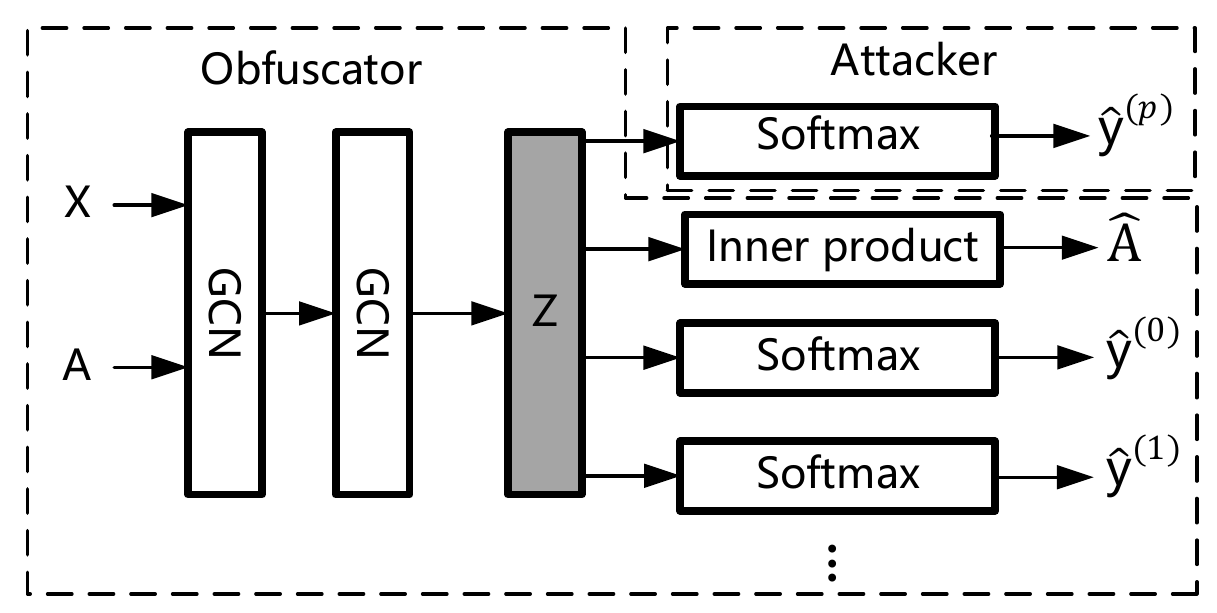}

 \caption{Adversarial Privacy-Purged Graph Embedding}
 \label{Fig:APPGE}
\end{figure}

\subsection{Adversarial Privacy-Purged Graph Embedding}
The adversarial training mechanism discussed above disentangles privacy latent factors from $\mathbf{Z}$ by disclosing the privacy labels directly as input to the decoder. We can also achieve a similar effect by disclosing the privacy labels as the output of the decoder. We call the latter mechanism Privacy-Purging and proposed  "Adversarial Privacy-Purged  Graph Embedding" (APPGE) as shown in Fig.~\ref{Fig:APPGE}.  APPGE consists of two networks (1) An Attacker  whose purpose is to extract privacy information from embedding matrix $\mathbf{Z}$; and (2) An Obfuscator which attempts to embed utility information and prevent the inference attacks launched by the attacker.

 Essentially, Attacker is a softmax classifier that aims to predict privacy labels $\mathbf{Y}^{(p)}$. We optimize the Attacker to minimize the following loss:
   \begin{equation}
   L_{att} =  -\frac{1}{|\mathbf{V}^{(p)}|}\sum_{i \in \mathbf{V}^{(p)}}\sum_{j=1}^{M_p}\mathbf{y}^{(p)}_{ij}\log\mathbf{\hat{y}}^{(p)}_{ij},
   \label{Eq:LossAttack}
   \end{equation}
 where $\mathbf{V}^{(p)}$ is the set of users on which the private attribute labels are available for training, $\mathbf{y}^{(p)}_{ij}$ and $\mathbf{\hat{y}}^{(p)}_{ij}$  are the $j$-th dimension value of $i$-th user's privacy one-hot label $\mathbf{y}^{(p)}_{i}$ and the   predicting result of privacy $\mathbf{\hat{y}}^{(p)}_{i}$   respectively, and $M_p$ is the dimension of private attribute.
   The architecture of Obfuscator is similar to GAE. The Obfuscator learns the latent code via GCN and decodes latent code to reconstruct network structure and utility attributes. In addition, it also tries to purge the latent code such that the attacker can't predict private attribute accurately. Therefore, we optimize the Obfuscator to minimize the reconstruction error and at the same time maximize the prediction error of the attacker. So the loss of the Obfuscator is as follows:
   \begin{equation}
   L_{obf} = L_{recon} - \lambda L_{att},
   \label{Eq:obf}
   \end{equation}
   where $\lambda$  is a hyperparameter that balances between utility and privacy. When we train the model, we update the Attacker and Obfuscator alternately until both modules' loss functions plateau.


\begin{figure*}[t]
 \centering
 \includegraphics[width=0.62\linewidth]{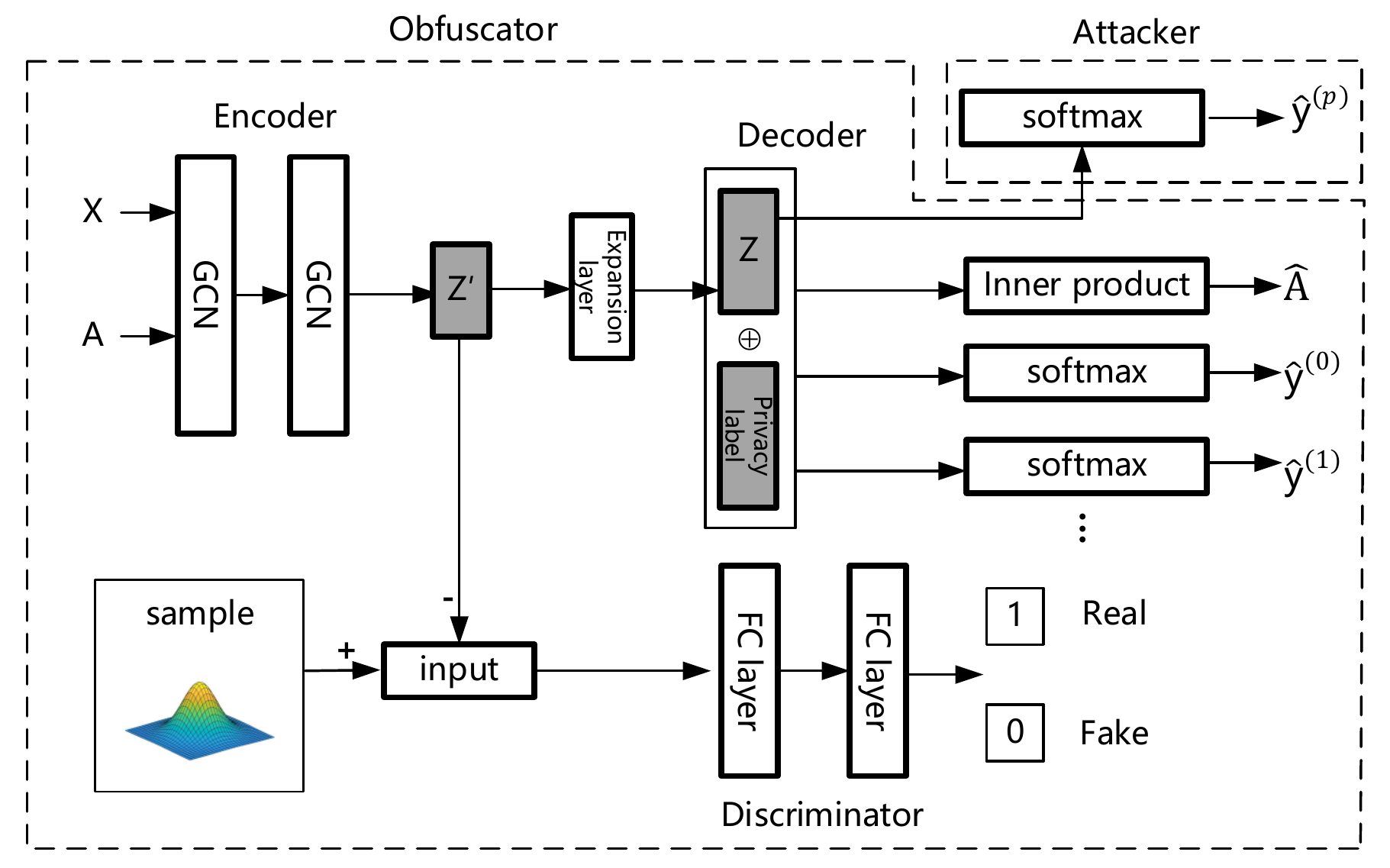}
 \caption{Adversarial Privacy Graph Embedding}
 \label{Fig:APGE}
 \end{figure*}
 \subsection{Adversarial Privacy Graph Embedding}
The Privacy-Disentangling and Privacy-Purging mechanisms we discussed above preserve the privacy information at the different stages of the training: the former is at the input of the decoder, while the latter at the output of the decoder, and they may work complementary to each other. Therefore, we integrate both mechanisms into one framework "Adversarial Privacy Graph Embedding" (APGE) to jointly protect users' privacy, with the model shown in Fig.~\ref{Fig:APGE}. As we can see, the architecture of APGE is very similar to APPGE that consists of an Obfuscator and an Attacker, while we integrate APDGE as the architecture of the Obfuscator. Note that the Attacker launches an inference attack on the latent code matrix $\mathbf{Z}$ rather than the latent code $\mathbf{Z^+}$, which is the concatenation of $\mathbf{Z}$ and privacy label encoding.

\begin{algorithm}[h]
  \caption{Adversarial Privacy Graph Embedding}
  \label{alg:APGE}
  \renewcommand{\algorithmicrequire}{ \textbf{Input:}}
 \renewcommand{\algorithmicensure}{ \textbf{Output:}}

  \begin{algorithmic}[1]
    \Require
      $\mathbf{G}(\mathbf{V},\mathbf{E},\mathbf{X})$:A graph with links and features;
      $T$: the number of iterations;
      $K_{dis}$: the number of steps for iterating Discriminator;
      $K_{att}$: the number of steps for iterating Attacker;
      $d$: the dimension of the latent variable.
    \Ensure
      $\mathbf{Z} \in \mathbb{R}^{n\times d}$;
  \For {iteration $= 1$ to $T$}
  \For{$k_1 = 1$ to $K_{att}$}
  \State Update the Attacker by minimizing the loss of Eq.\ref{Eq:LossAttack}
  \EndFor
  \State Update the Encoder and Decoder to minimize the loss of Eq.\ref{Eq:obf}
  \For{$k_2 = 1$ to $K_{dis}$}
  \State Sample from  Gaussian distribution and $\mathbf{Z'}$
  \State Update the Discriminator to distinguish  the  real  samples from the fake samples by minimizing the loss of  Eq.\ref{Eq:LossDc}
  \EndFor
  \State Update the Encoder to fool the discriminator by maximizing the loss Eq.\ref{Eq:LossDc}
  \EndFor \\
  \Return {$\mathbf{Z}$}
  \end{algorithmic}
\end{algorithm}

We present the training process of APGE in the Algorithm 1. Given a garph $\mathbf{G}$, the Attacker and Obfuscator of APGE can be optimized by alternating gradient descent in two stages. Firstly, for the Attacker, we minimize the prediction error of the private attribute by  Eq.\ref{Eq:LossAttack} as shown in Line 2-4. Secondly, for the Obfuscator, whose training process is similar to ADPGE, we update the model as shown in Line 5-10. After finishing the training, we get the privacy-preserved graph embedding matrix $\mathbf{Z}$.

\begin{table*}
\small
      \caption{Utility and Privacy Evaluation on Yale}
      \label{Tab:Yale}
 \adjustbox{center}{
      \begin{tabular}{lcccccccc}
      \hline
               Method & \multicolumn{2}{c}{link} & \multicolumn{2}{c}{utility attributes(status)} &\multicolumn{4}{c}{private attribute(class year)} \\
        \hline
           &\multirow{2}{*}{ACC}  &\multirow{2}{*}{Macro F1}     &\multirow{2}{*}{ACC} &\multirow{2}{*}{Macro F1}  &\multicolumn{2}{c}{MLP} &\multicolumn{2}{c}{SVM}  \\
           \cline{6-7} \cline{8-9}\
           &&&&   & ACC  &Macro F1 & ACC  &Macro F1\\

        \hline
       GAE     &0.856 \scriptsize{$\pm$ 0.003 }&0.856 \scriptsize{$\pm$ 0.003}&0.923 \scriptsize{$\pm$ 0.002}&0.927 \scriptsize{$\pm$ 0.002}&0.931\scriptsize{$\pm$ 0.002}&0.930 \scriptsize{$\pm$ 0.002}& 0.940 \scriptsize{$\pm$ 0.001 }&0.941 \scriptsize{$\pm $0.001}\\
       GAE\_RM  &0.854  \scriptsize{$\pm$ 0.002 }&0.854 \scriptsize{$\pm$ 0.002 }&0.933 \scriptsize{$\pm$ 0.003 }&0.930\scriptsize{$\pm$ 0.003 }&0.907\scriptsize{$\pm$ 0.002  }&0.906\scriptsize{$\pm$ 0.002 }&0.918\scriptsize{$\pm$ 0.001 }&0.918 \scriptsize{$\pm$ 0.001 }\\
        CDSPIA  &0.824 \scriptsize{$\pm$ 0.002 }&0.824 \scriptsize{$\pm$ 0.002 }&0.920 \scriptsize{$\pm$ 0.001 }&0.916 \scriptsize{$\pm$ 0.001 }&0.896\scriptsize{$\pm$ 0.003 }&0.896\scriptsize{$\pm$ 0.003 }&0.894\scriptsize{$\pm$ 0.001 }&0.893 \scriptsize{$\pm$ 0.001}\\
        APDGE     &0.840  \scriptsize{$\pm$ 0.001 }&0.840  \scriptsize{$\pm$ 0.001 }&0.884 \scriptsize{$\pm$ 0.003 }&0.877\scriptsize{$\pm$ 0.003  }&0.824\scriptsize{$\pm$ 0.009 }&0.825\scriptsize{$\pm$ 0.009 }&0.798\scriptsize{$\pm$ 0.008 }&0.801 \scriptsize{$\pm$ 0.007} \\
        APPGE     &0.856 \scriptsize{$\pm$ 0.002 }&0.856 \scriptsize{$\pm$ 0.002 }&0.929 \scriptsize{$\pm$ 0.005 }&0.923 \scriptsize{$\pm$ 0.005  }&0.890\scriptsize{$\pm$ 0.011 }&0.890\scriptsize{$\pm$ 0.011 }&0.891\scriptsize{$\pm$ 0.009 }&0.890 \scriptsize{$\pm$ 0.009} \\
        APGE     &0.820 \scriptsize{$\pm$ 0.002 }&0.820 \scriptsize{$\pm$  0.002 }&0.813 \scriptsize{$\pm$  0.010 }&0.780 \scriptsize{$\pm$ 0.009 }&\textbf{0.568 \scriptsize{$\pm$ 0.014} }&\textbf{0.579 \scriptsize{$\pm$ 0.016} }&\textbf{0.514 \scriptsize{$\pm$ 0.013} }&\textbf{0.518 \scriptsize{$\pm$ 0.016}}\\
       \hline
      \end{tabular}}
    \end{table*}

\begin{table*}

    \small
      \caption{Utility and Privacy Evaluation on Rochester}
      \label{Tab:Rochester}

      \adjustbox{center}{
      \begin{tabular}{lcccccccc}
        \hline
         Method & \multicolumn{2}{c}{link} & \multicolumn{2}{c}{utility attributes(class year)} &\multicolumn{4}{c}{private attribute(gender)}                  \\
        \hline
           &\multirow{2}{*}{ACC}  &\multirow{2}{*}{Macro F1}     &\multirow{2}{*}{ACC} &\multirow{2}{*}{Macro F1}  &\multicolumn{2}{c}{MLP} &\multicolumn{2}{c}{SVM}  \\
           \cline{6-7} \cline{8-9}
           &&&&& ACC  &Macro F1   & ACC  &Macro F1 \\
           \hline

      GAE      &0.862 \scriptsize{$\pm$ 0.001 }&0.862 \scriptsize{$\pm$ 0.001 }&0.953 \scriptsize{$\pm$  0.002 }&0.947 \scriptsize{$\pm$ 0.002  }&0.781\scriptsize{$\pm$  0.003 }&0.781 \scriptsize{$\pm$ 0.003 }&0.769 \scriptsize{$\pm$ 0.002  }&0.768 \scriptsize{$\pm$ 0.002}\\
        GAE\_RM   &0.857  \scriptsize{$\pm$ 0.001 }&0.857 \scriptsize{$\pm$ 0.002 }&0.948\scriptsize{$\pm$ 0.002 }&0.941\scriptsize{$\pm$  0.002 }&0.732\scriptsize{$\pm$ 0.002  }&0.731\scriptsize{$\pm$ 0.002 }&0.724\scriptsize{$\pm$ 0.001 }&0.722 \scriptsize{$\pm$ 0.001}\\
        CDSPIA   &0.848 \scriptsize{$\pm$  0.002 }&0.848 \scriptsize{$\pm$ 0.002  }&0.862 \scriptsize{$\pm$ 0.002 }&0.851 \scriptsize{$\pm$  0.002  }&0.666\scriptsize{$\pm$ 0.002 }&0.666\scriptsize{$\pm$ 0.002 }&0.655\scriptsize{$\pm$ 0.001 }&0.654 \scriptsize{$\pm$  0.001}\\
        APDGE      &0.849  \scriptsize{$\pm$  0.002 }&0.849 \scriptsize{$\pm$ 0.002 }&0.924 \scriptsize{$\pm$0.003  }&0.915 \scriptsize{$\pm$ 0.003  }&0.654\scriptsize{$\pm$ 0.003 }&0.651\scriptsize{$\pm$ 0.003 }&0.655\scriptsize{$\pm$  0.004 }&0.652 \scriptsize{$\pm$ 0.004 }\\
        APPGE     &0.857 \scriptsize{$\pm$ 0.002  }&0.857 \scriptsize{$\pm$  0.002 }&0.895 \scriptsize{$\pm$ 0.003 }&0.889 \scriptsize{$\pm$ 0.003   }&0.717\scriptsize{$\pm$ 0.004 }&0.717\scriptsize{$\pm$  0.004 }&0.717\scriptsize{$\pm$ 0.003 }&0.717 \scriptsize{$\pm$  0.003}\\

        APGE     &0.848 \scriptsize{$\pm$ 0.002  }&0.848 \scriptsize{$\pm$ 0.002  }&0.923 \scriptsize{$\pm$ 0.005  }&0.914 \scriptsize{$\pm$ 0.005  }&\textbf{0.636 \scriptsize{$\pm$ 0.004} }&\textbf{0.634 \scriptsize{$\pm$ 0.004} }&\textbf{0.649 \scriptsize{$\pm$ 0.007} }&\textbf{0.645 \scriptsize{$\pm$ 0.007} }\\
       \hline

      \end{tabular}
      }
    \end{table*}

\section{Experiments}
In this section, we empirically demonstrate the effectiveness of our method,APGE,
and compare it with the start-of-the-art privacy preserving method.
We also evaluate the model's robustness to different attack models and the impacts of the hyperparameters and expansion layer in our method.
\subsection{Dataset and Baseline Methods}
We conduct experiments on two real-world graph datasets: \textbf{Yale} and \textbf{Rochester}~\footnote{https://escience.rpi.edu/data/DA/fb100/} which collect all the facebook friendships of Yale University and Rochester University in 2005 as well as a number of user attributes including student/faculty status (status for short), gender, major, second major, class year, dorm/house, and high school. The two networks contain 8578 nodes, 405450 edges and 4563 nodes, 167653 edges, respectively. For Yale, class year (6 categories) is set as the sensitive attribute and status is set as the utility attribute; while for Rochester, we regard gender (2 categories) as a sensitive attribute and class year as the utility attribute. Indeed, if the labels of sensitive attribute are provided, our methods can also deal with the cases where the other attributes are private, {\em e.g.,} sexual orientation and political opinion.

We evaluate the performance of proposed models against the following approaches: (1) \textbf{GAE}~\cite{kipf2016variational} learns network representation by autoencoder where encoder is graph convolutional network and decoder consists of inner product to reconstruct adjacency matrix and softmax function to predict utility attributes. (2) \textbf{GAE\_RM} ~\cite{kipf2016variational} is a framework similar to that of GAE. But the sensitive attributes are removed from the input attribute matrix $\mathbf{X}$ directly. (3) \textbf{CDSPIA}~\cite{cai2018collective} is a state-of-the-art approach to defending inference attacks on graph by deleting or perturbing users' attributes and linkages which are closely related to privacy. We use CDSPIA to sanitize the graph dataset and then embed the graph data processed via GAE.

All the methods embed the graphs to a 64-dimensional space. Particularly, in both APDGE and APGE, we first compress each node's information to a 16-dimensional vector $\mathbf{z}'_i$ for Yale and a 8-dimensional vector for Rochester. For APPGE and APGE, the $\lambda$ in Eq.~\ref{Eq:obf} is set as 1 and 10 for Yale and Rochester, respectively

\subsection{Utility and Privacy Evaluation} \label{sec:UPeva}
In this part of experiments, we verify the effectiveness of the three models we proposed on utility reservation and privacy protection and compare them  with the baselines. We qualify the utility in term of prediction accuracy of link and utility attribute from the embeddings. For link prediction, we compare models based on the classifier's ability to correctly classify edges and non-edges. Here the classifier is multi-layer perceptron (MLP)~\cite{pal1992multilayer} trained on embeddings.  
For utility attribute prediction, an MLP is utilized to classify the utility attributes based embeddings. Meanwhile, we qualify the privacy in term of prediction accuracy of sensitive attribute. Both MLP and support vector machine (SVM)~\cite{cortes1995support} are employed as the adversaries to predict the sensitive attribute. For each models, we repeat the experiment 10 times and report the average results and standard deviation in Table \ref{Tab:Yale} and Table \ref{Tab:Rochester}.

Comparing the performance of GAE and GAE\_RM, we can see that even if the private attribute is removed from the input of embedding model, the attacker is still able to infer the user's privacy from embeddings by the inherent relation between the private attribute and the input of embedding model. It can be seen from Table \ref{Tab:Yale} that APDGE obtains an average Macro-F1 of 0.813 on private attribute, outperforming the previous approach CDSPIA. Meanwhile, we observe that APDGE and CDSPIA have similar performances on preserving utilities. This indicates that the end to end training of graph embedding and privacy protection provides a stronger system than the decoupled method that processes the two steps separately for preserving privacy in graph embedding.

From the results on both the Yale and Rochester datasets, we find that APGE demonstrates the strongest capability of protecting user's privacy while retaining the utilities. On Yale, APGE achieves an average Macro-F1 of 0.549 on private attribute, outperforming APDGE by a significant margin of 30\%. At the same time, the accuracies of link and utility attribute are preserved largely. This demonstrates that the integration of disentangling and purging mechanisms brings an additional and significant gain in performance. We observe that the preserving privacy performance of APGE is more significant on Yale than on Rochester. We note that this is because the privacy is more closely correlated with the utility in the Yale dataset than Rochester. That is, user's friendship and status are strongly related to her class year. At the same time, user's linkage and class year are almost irrelevant to her gender.

\begin{figure}[!ht]
\centering
     \subfloat[Yale\label{Classcial AAE}]{%
       \includegraphics[width=0.245\textwidth]{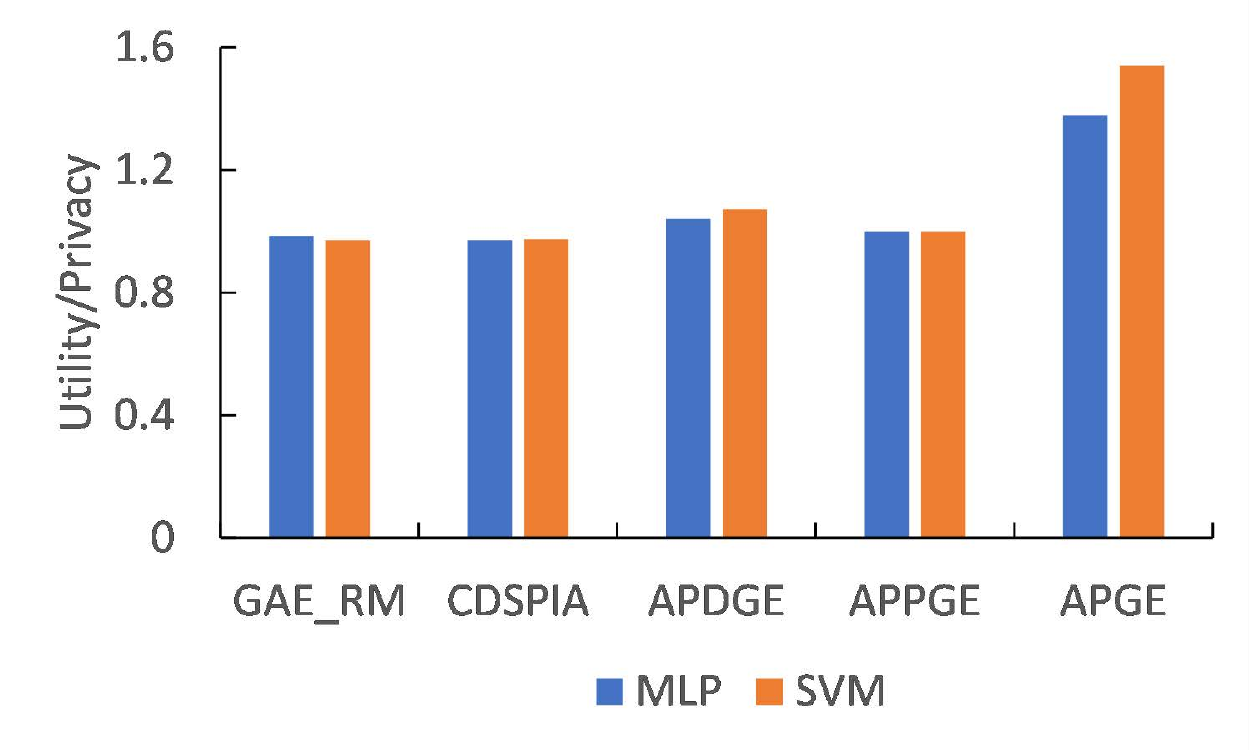}
     }
     \subfloat[Rochester\label{Supervised AAE}]{%
       \includegraphics[width=0.245\textwidth]{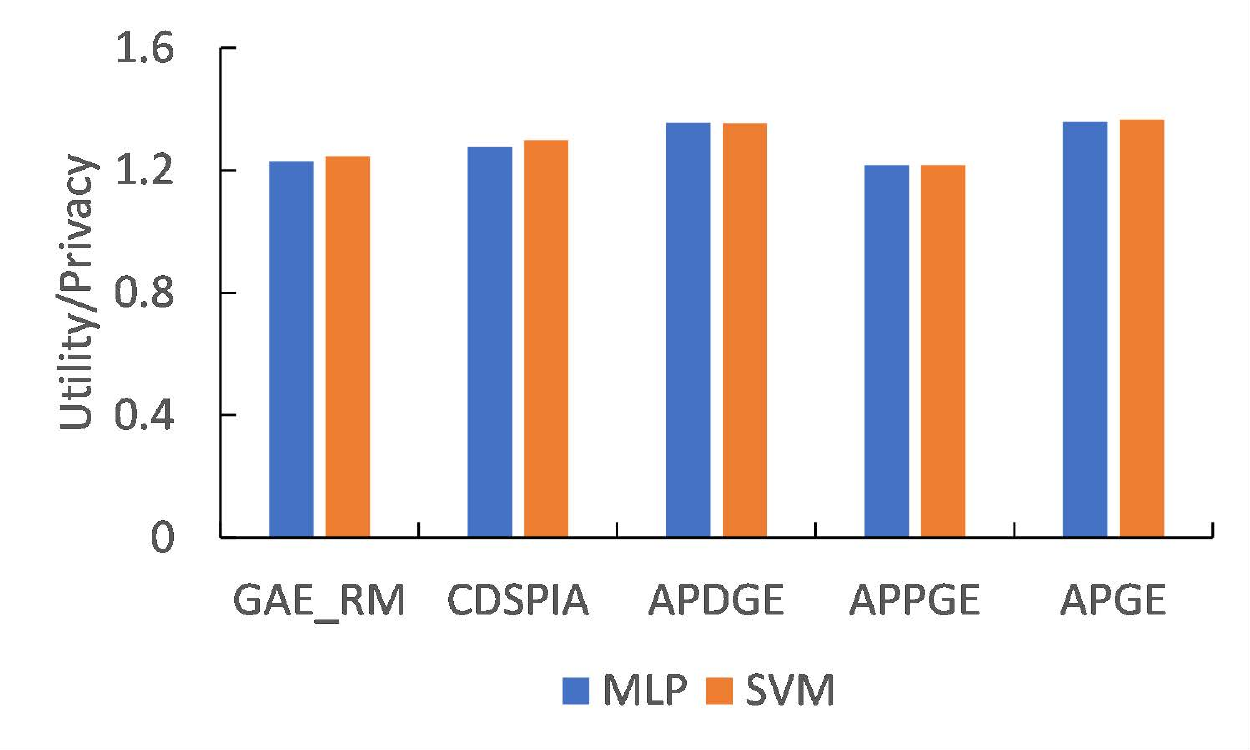}
     }
     \caption{Evaluation on the trade off between utility and privacy}
     \label{fig:u_p}
   \end{figure}

Moreover, to illustrate the trade off between utility and privacy, Fig.~\ref{fig:u_p} reports the ratio of utility (that is the average Macro-F1 of link and utility attributes) to privacy ({\em i.e.}, the  Macro-F1 of private attribute). As we can see, APGE achieves the best performance when both utility preservation and privacy protection are taken into account.

\begin{figure}[!ht]
\centering
     \subfloat[Impact on utility\label{subfig-1:dummy}]{%
       \includegraphics[width=0.24\textwidth]{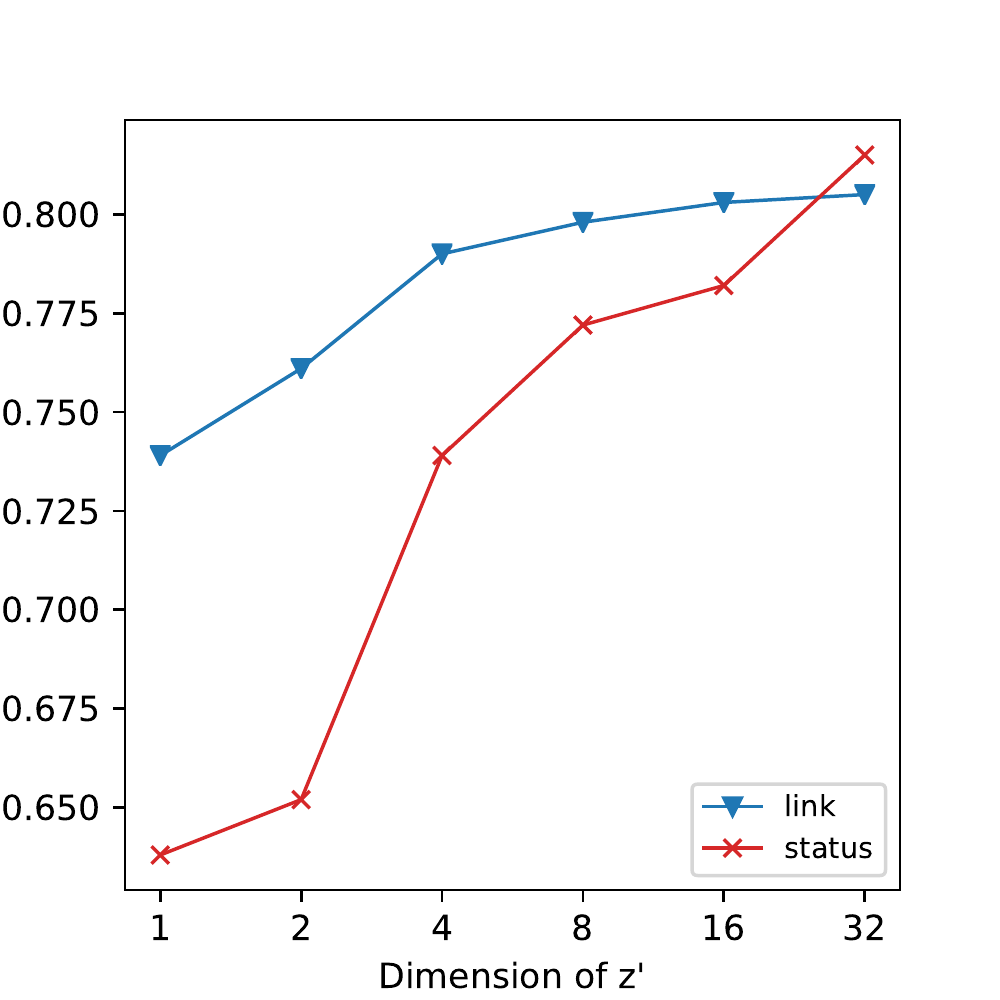}
     }
     \subfloat[Impact on privacy\label{subfig-2:dummy}]{%
       \includegraphics[width=0.24\textwidth]{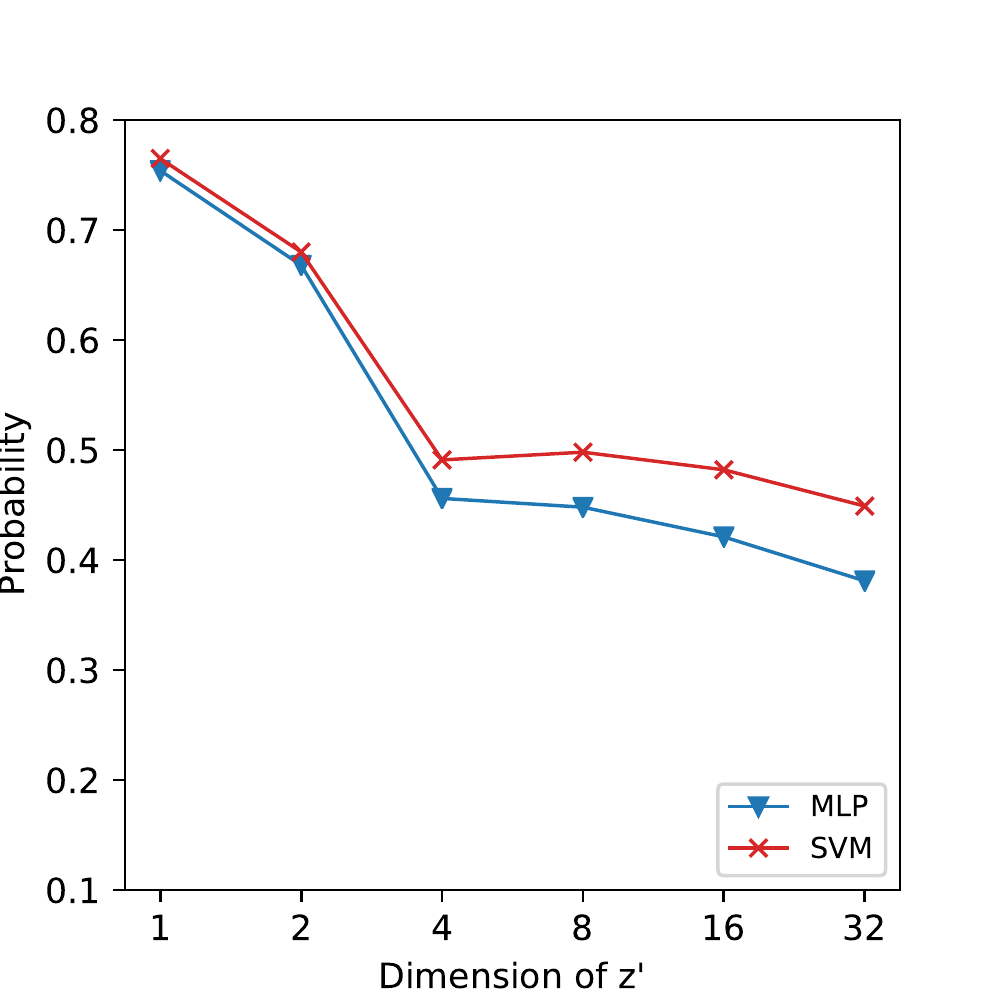}
     }
     \caption{Impact of the hidden code's dimension }
     \label{fig:dimensionImpact}
\end{figure}

\begin{figure}[!ht]
\centering
     \subfloat[Impact on utility\label{subfig-1:dummy}]{%
       \includegraphics[width=0.24\textwidth]{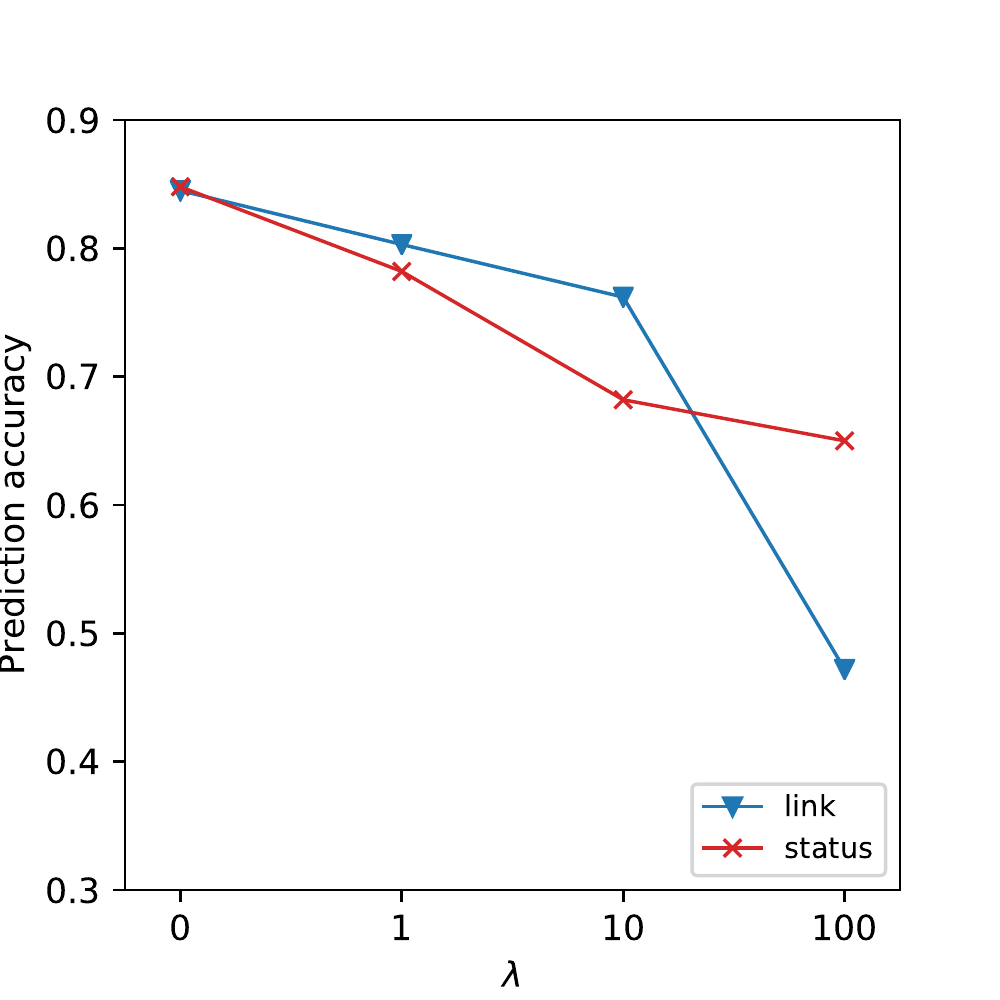}
     }
     \subfloat[Impact on privacy\label{subfig-2:dummy}]{%
       \includegraphics[width=0.24\textwidth]{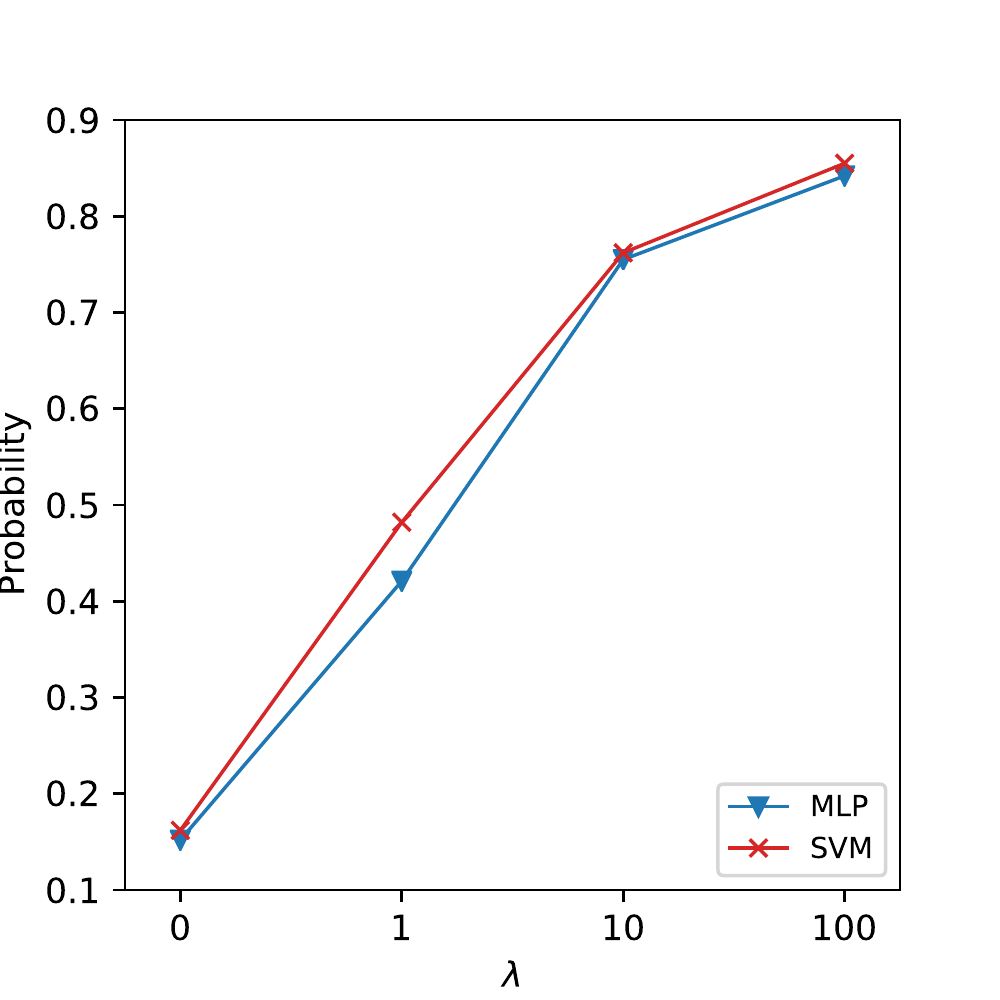}
     }
     \caption{Impact of the $\lambda$ }
     \label{fig:lambdaImpact}

\end{figure}
\subsection{Impact of hyperparameters}\label{sec:hyperParameters}
The dimension of hidden code  $\mathbf{z}'_i$  and the hyperparameter $\lambda$ have significant impacts to the performance of APGE. Fig.~\ref{fig:dimensionImpact} shows the performance of APGE on privacy pretection and utility preserving on Yale when the hidden code dimension increases. Here, we define ``1 - prediction accuracy of private attribute" as  the probability of successfully preventing inference attack. As we can see, when the dimension of hidden code increases, link and utility attributes can be recovered more accurately. On the other hand, the higher dimension of hidden code is, the more private information is retained, leading to a lower probability of preventing inference attack. Similarly, Fig.~\ref{fig:lambdaImpact} shows the performance of APGE on Yale when $\lambda$ increases from 0 to 100. As expected, when $\lambda$ increases, we observe a monotonic decrease of utility prediction accuracy and a monotonic increase of the probability of resisting inference attack. These results demonstrate that we could adjust the dimension of $\mathbf{z}'_i$  and $\lambda$ to achieve a desired utility-privacy tradeoff.

\subsection{Robustness against attack models}
\begin{figure}[t]
\centering
     \subfloat[MLP]{%
       \includegraphics[width=0.245\textwidth]{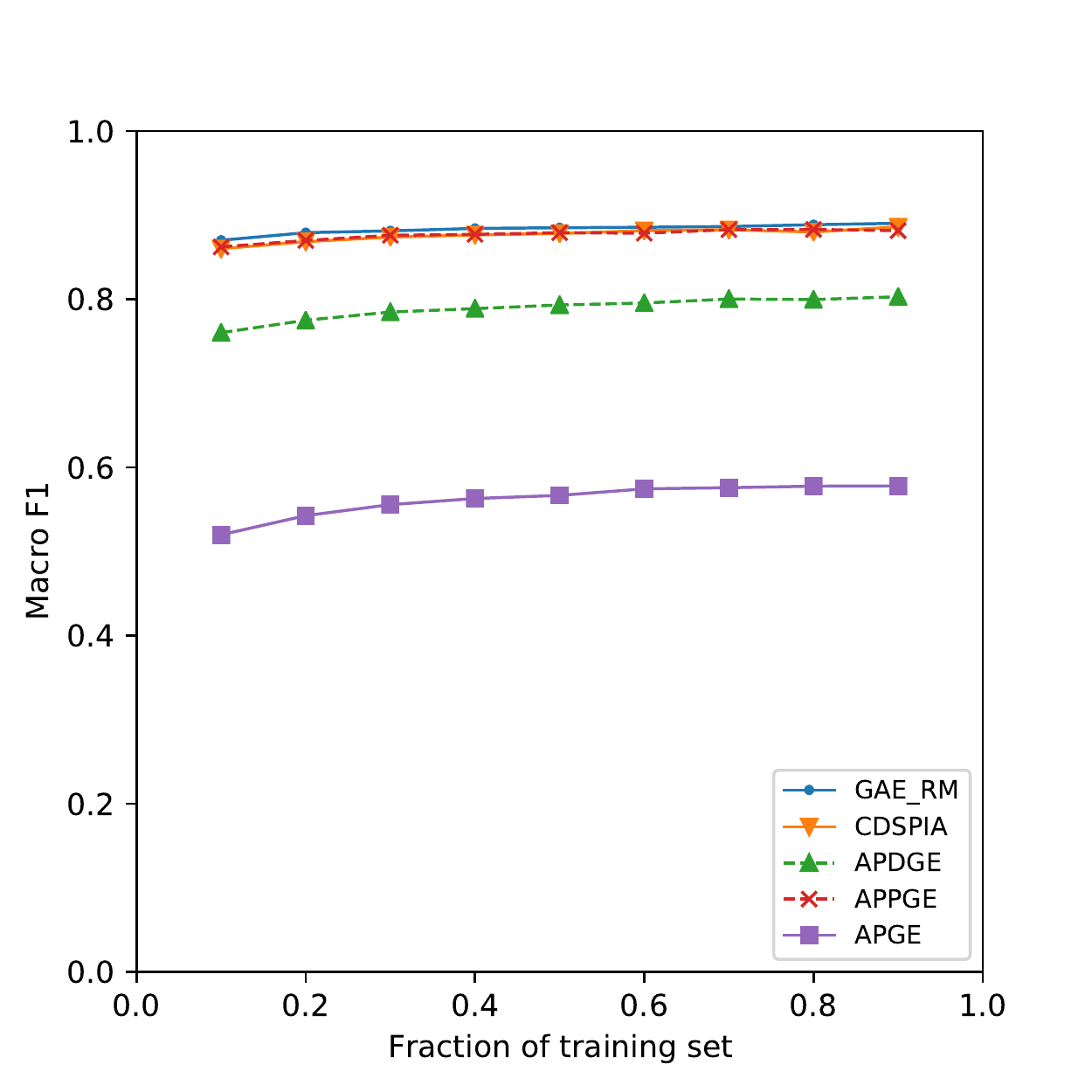}
     }
     \subfloat[SVM]{%
       \includegraphics[width=0.245\textwidth]{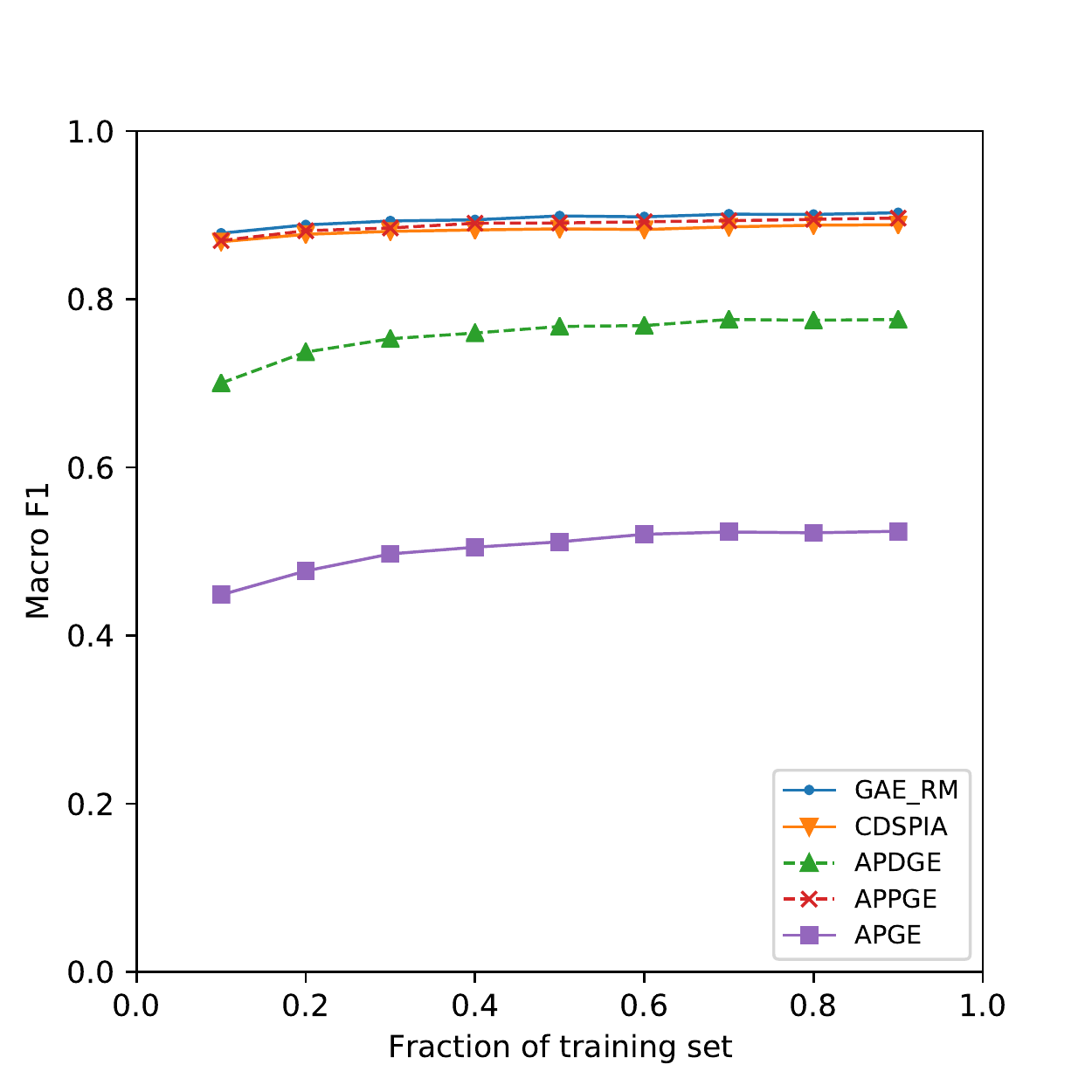}
     }\\
      \subfloat[KNN]{%
       \includegraphics[width=0.245\textwidth]{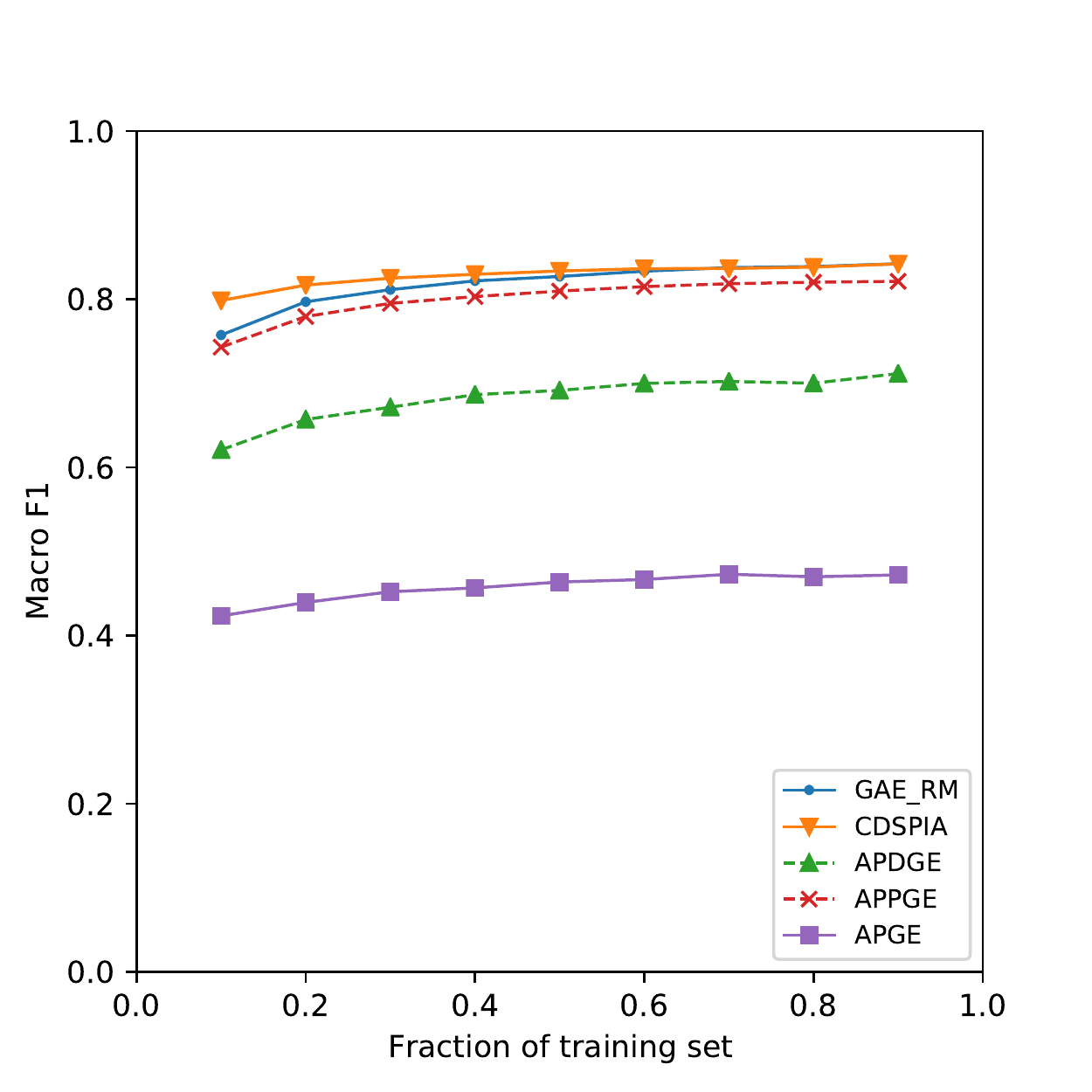}
     }
     \subfloat[Random Forest]{%
       \includegraphics[width=0.245\textwidth]{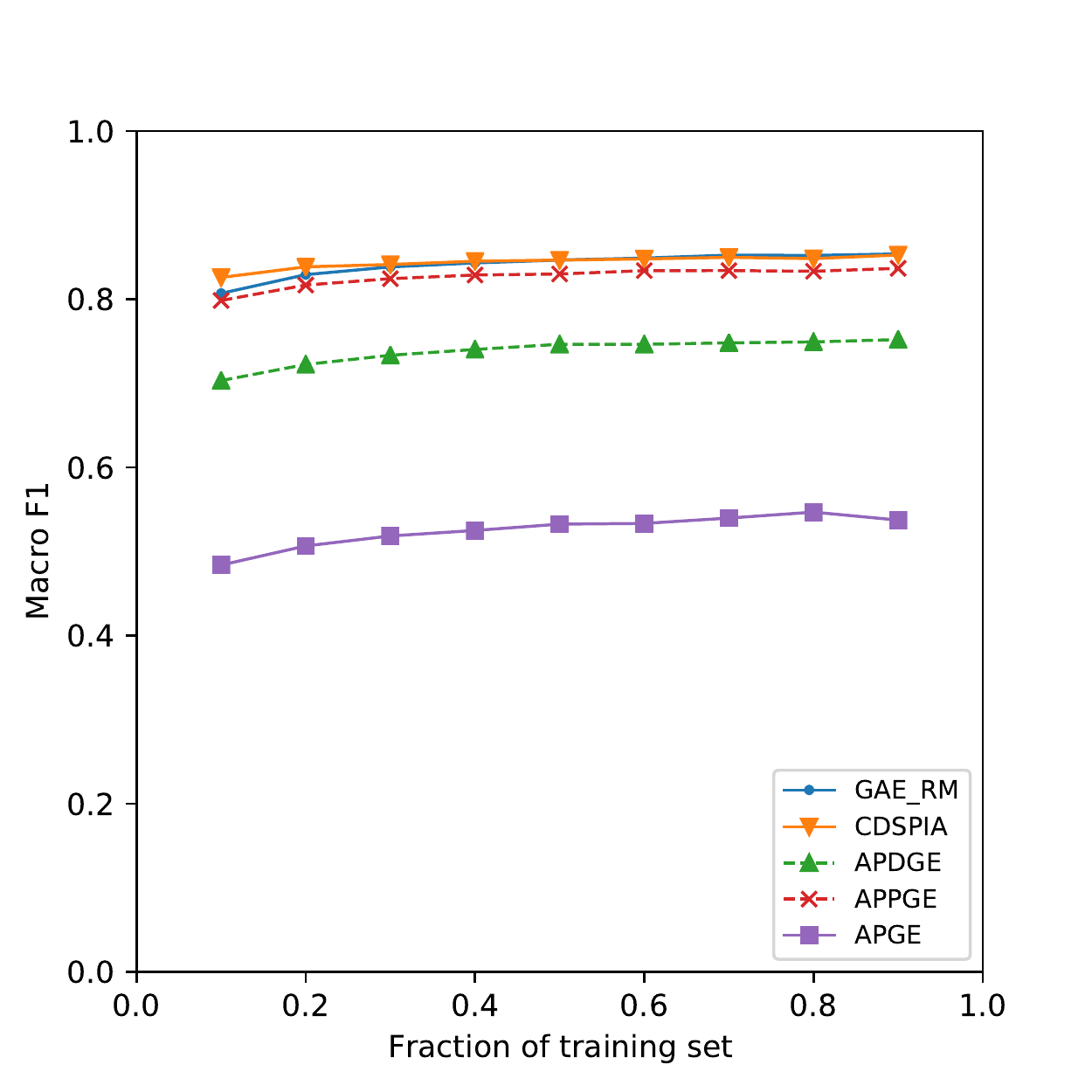}
     }\\
      \subfloat[Adaboost]{%
       \includegraphics[width=0.245\textwidth]{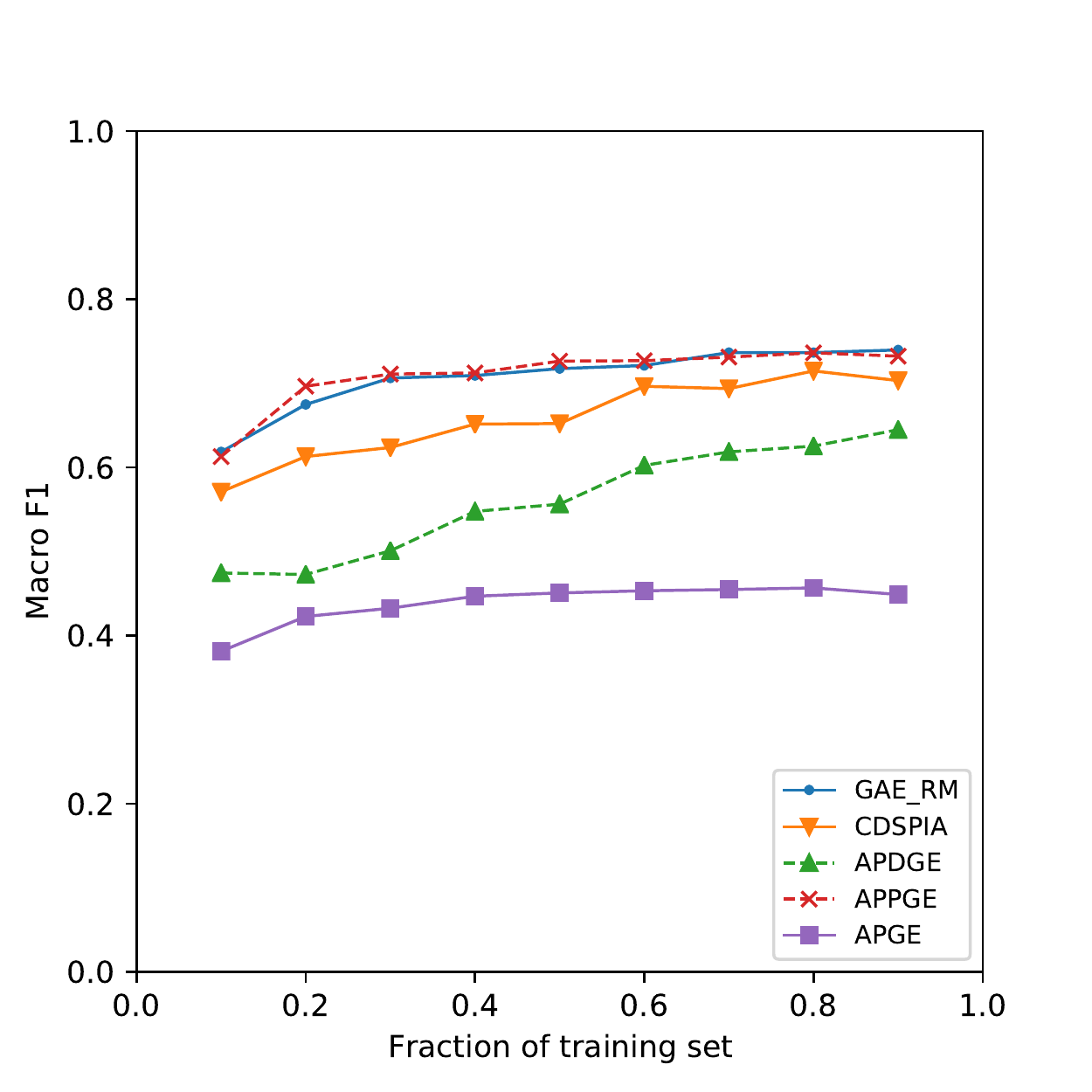}
     }
     \subfloat[Gradient boosting]{%
       \includegraphics[width=0.245\textwidth]{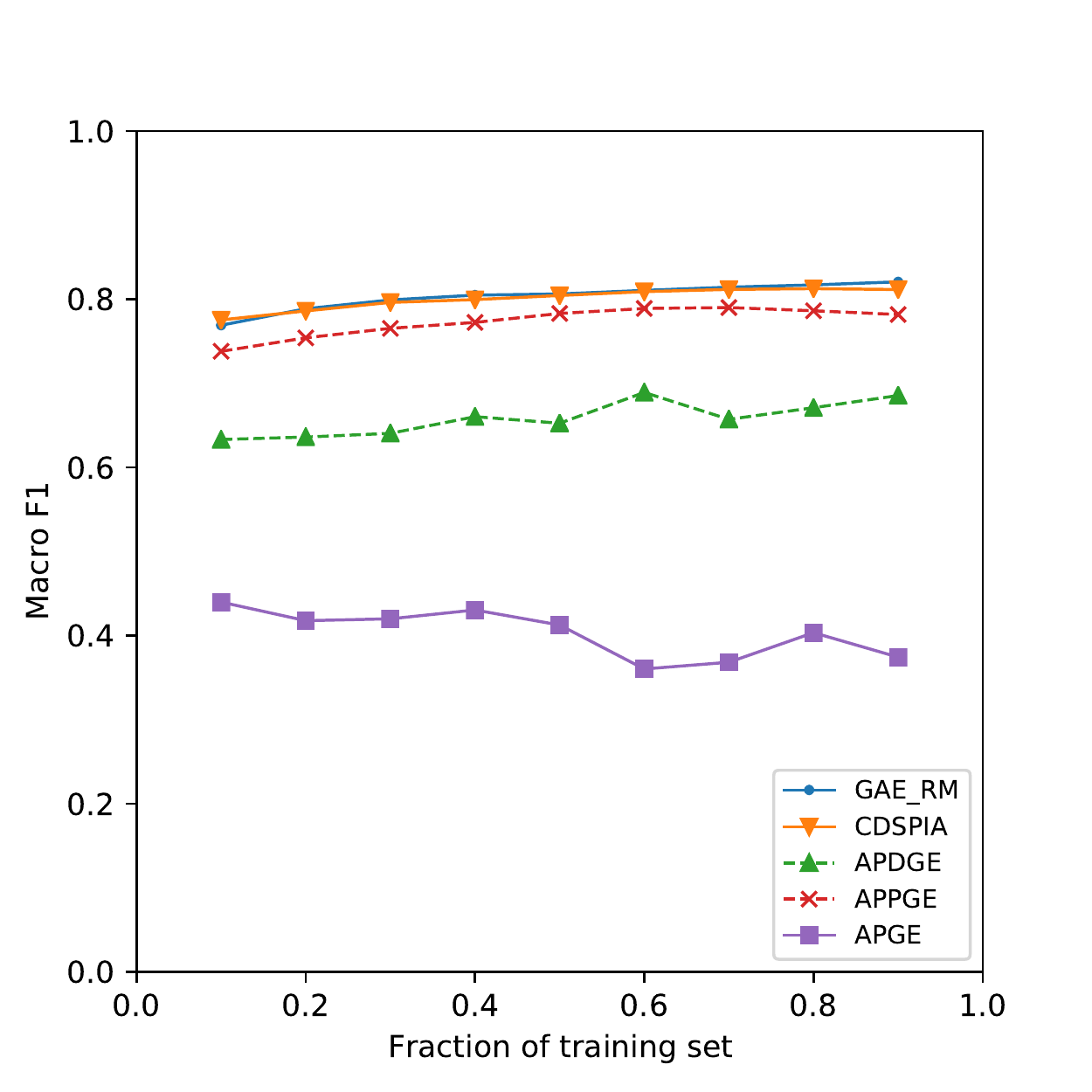}
     }
     \caption{Inference attack on Yale with different attack models}
     \label{fig:DiffAttackYale}
\end{figure}

\begin{figure}[t]
\centering
     \subfloat[MLP]{%
      \includegraphics[width=0.245\textwidth]{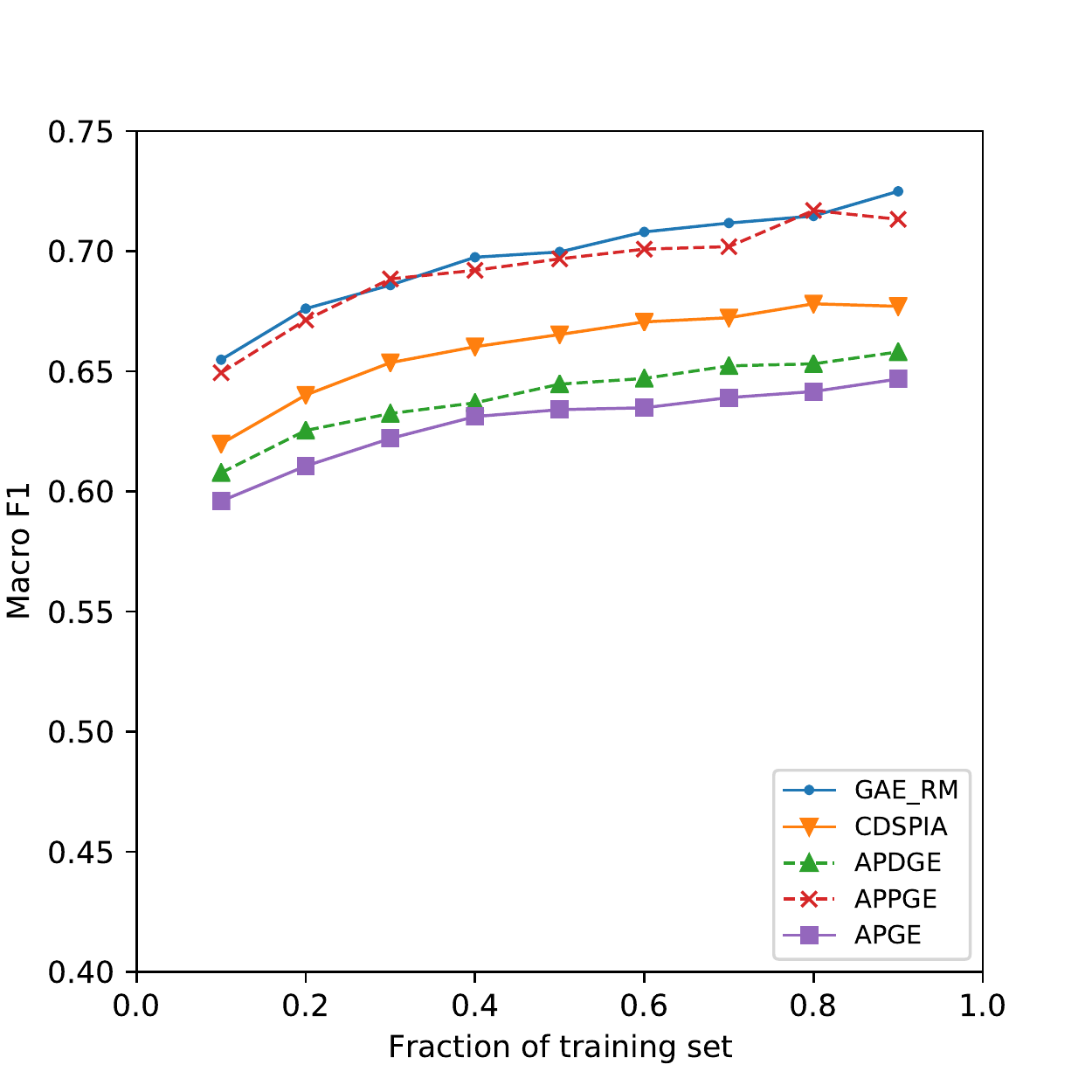}
     }
     \subfloat[SVM]{%
      \includegraphics[width=0.245\textwidth]{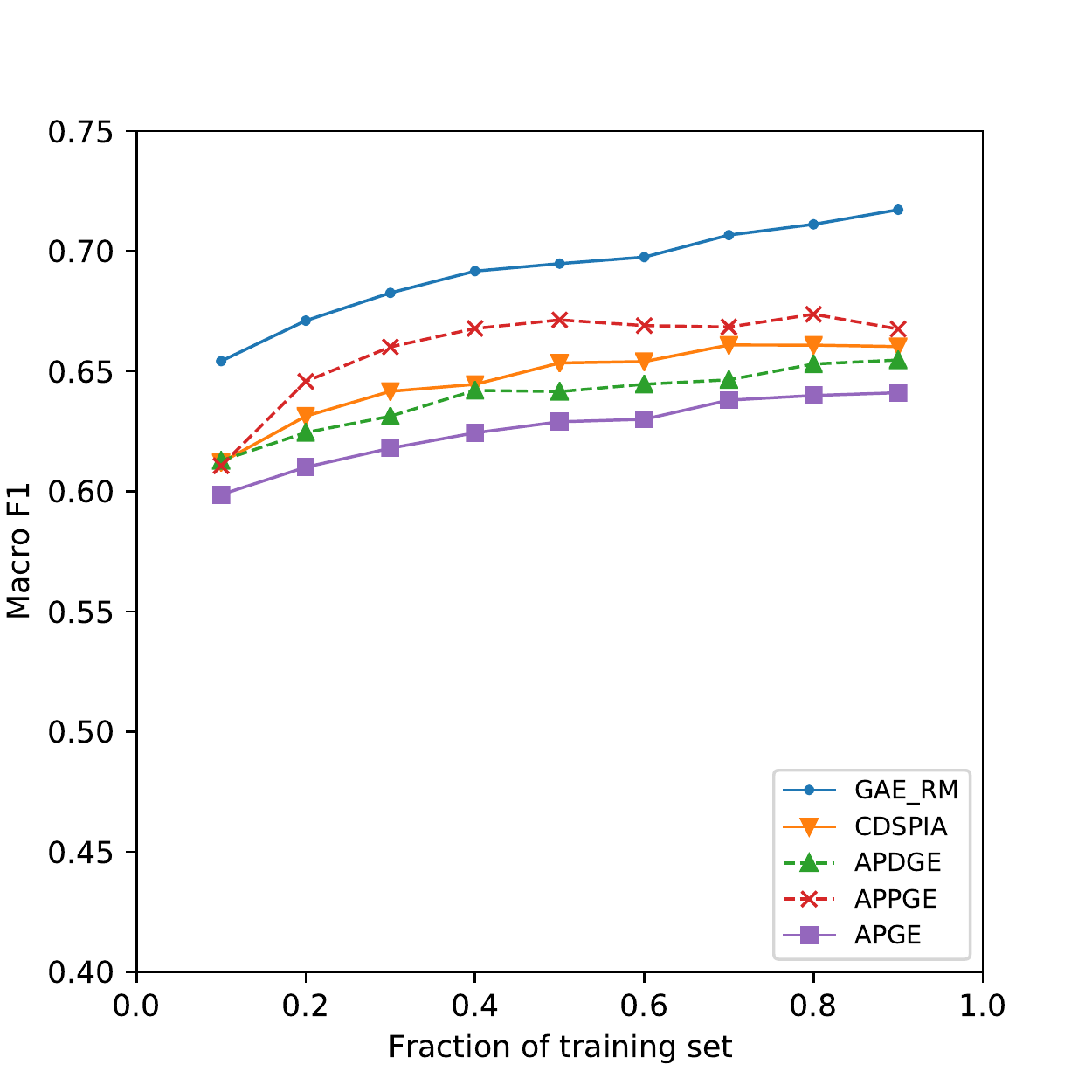}
     }\\
      \subfloat[KNN]{%
      \includegraphics[width=0.245\textwidth]{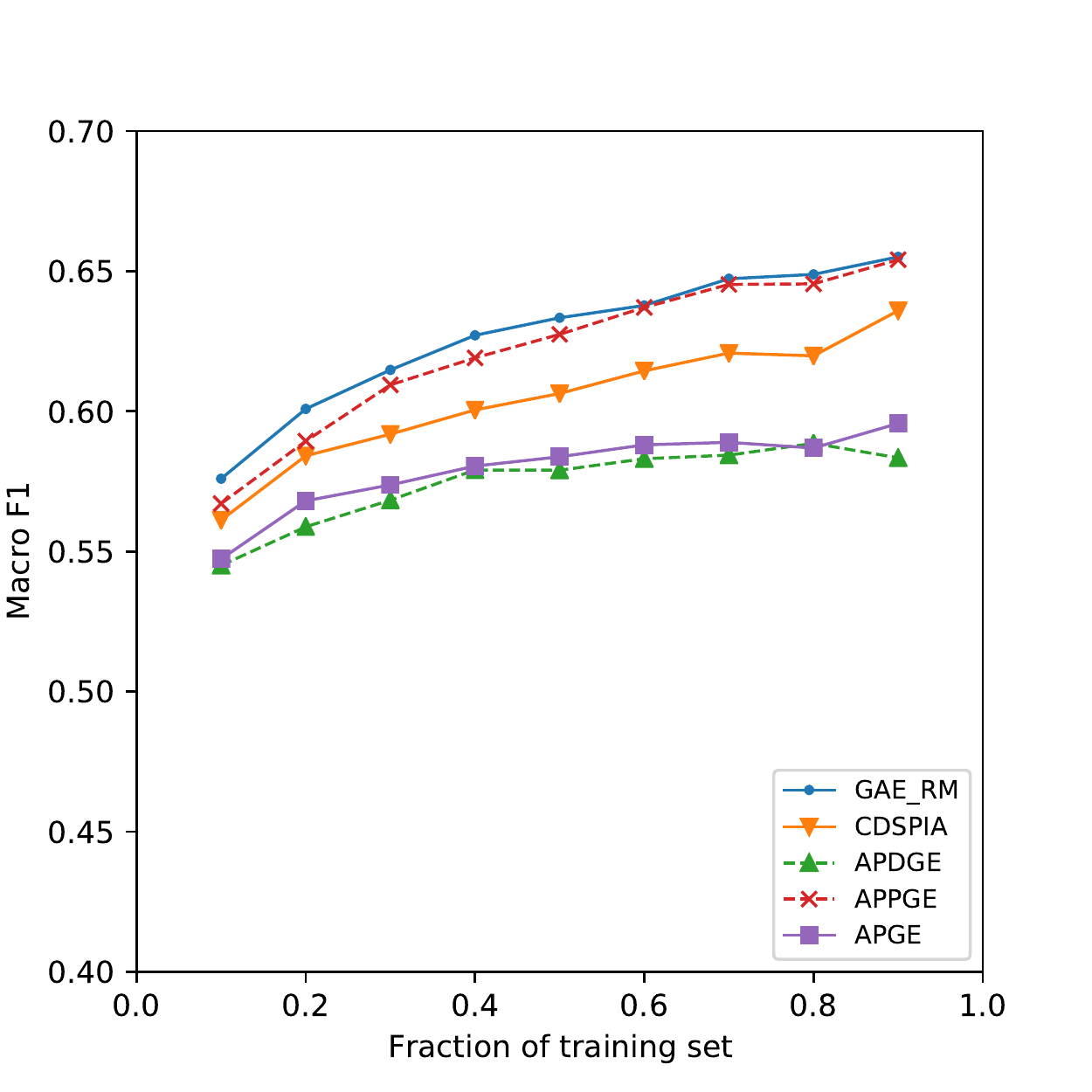}
     }
     \subfloat[Random Forest]{%
      \includegraphics[width=0.245\textwidth]{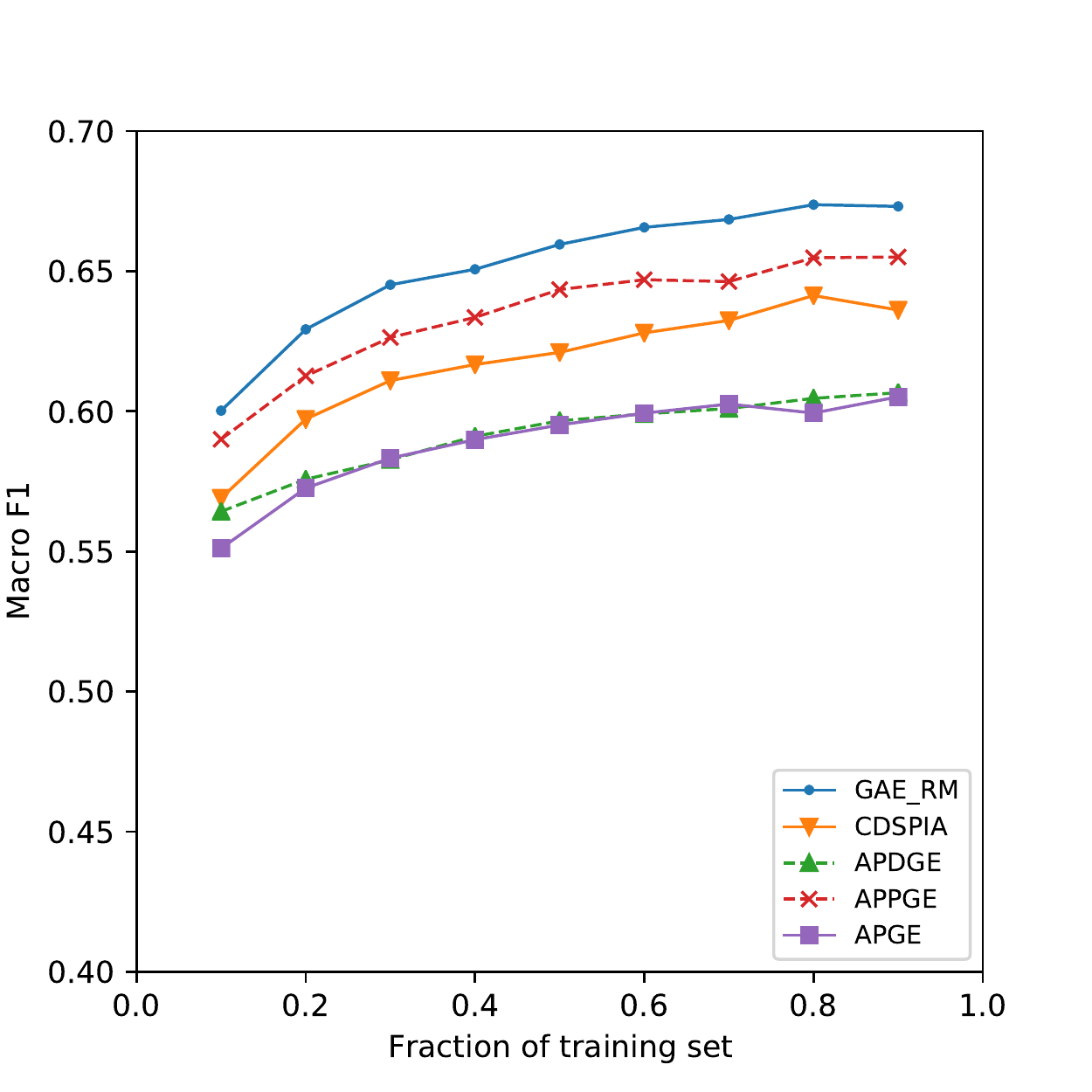}
     }\\
      \subfloat[Adaboost]{%
      \includegraphics[width=0.245\textwidth]{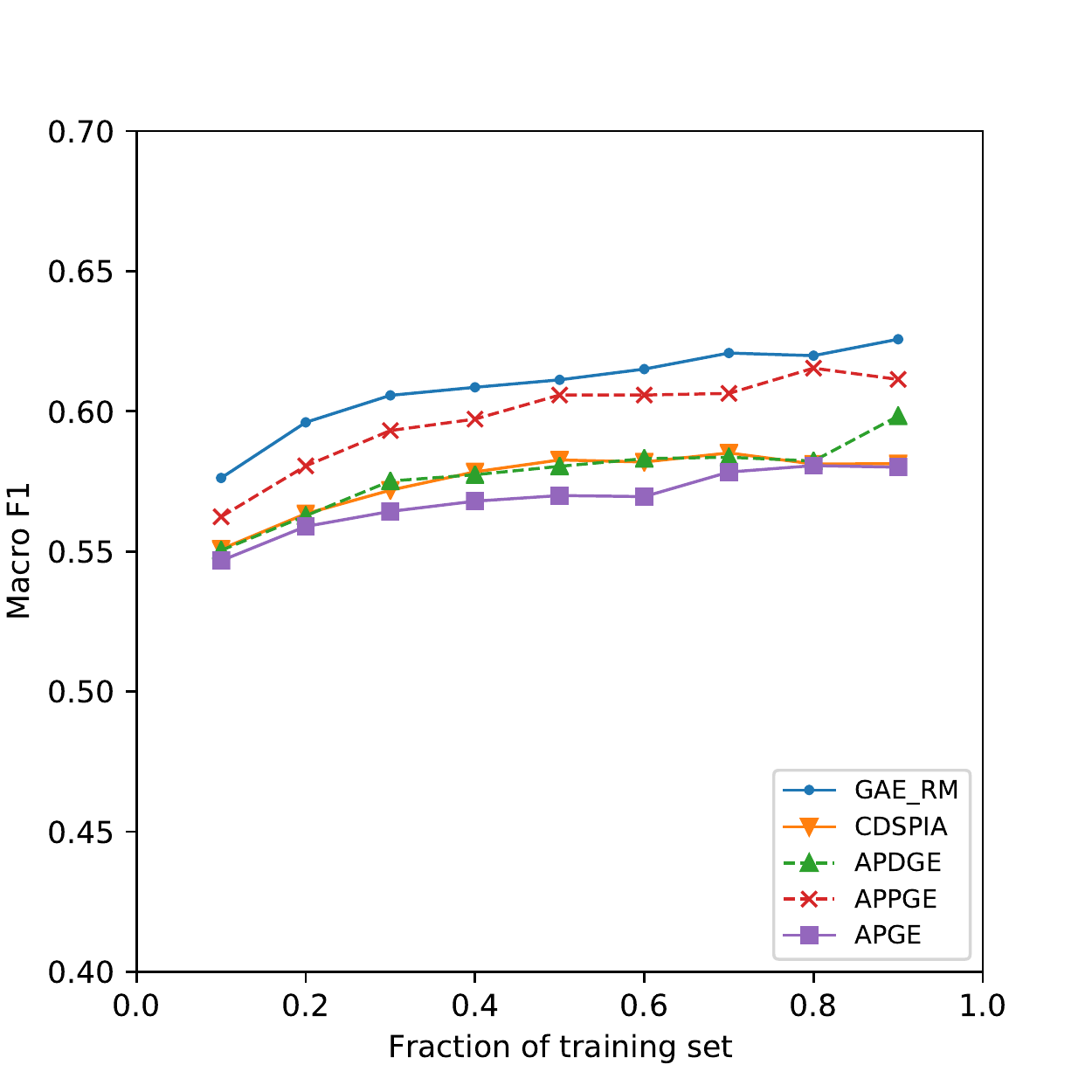}
     }
     \subfloat[Gradient boosting]{%
      \includegraphics[width=0.245\textwidth]{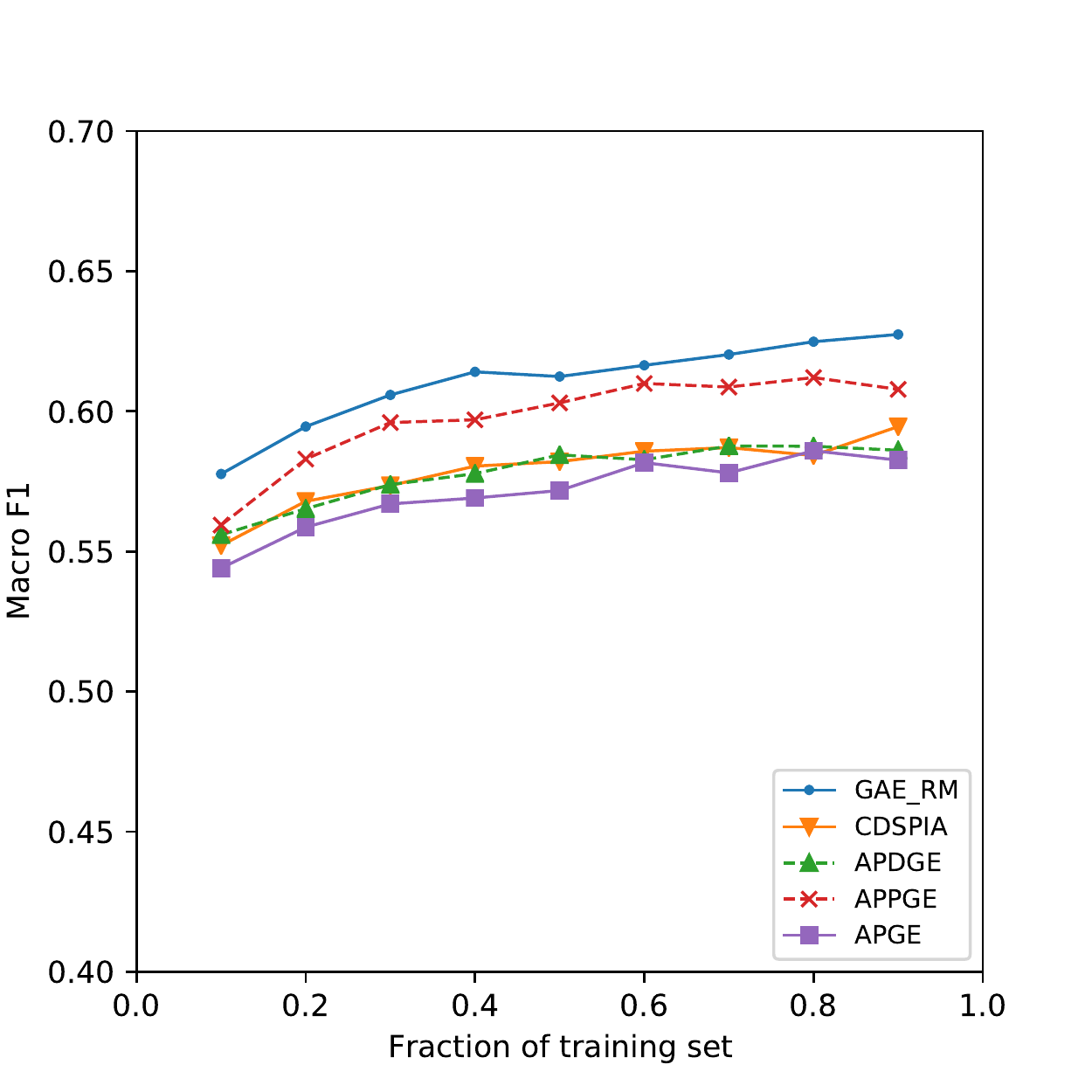}
     }
     \caption{Inference attack on Rochester with different attack models}
     \label{fig:DiffAttackRochester}
\end{figure}

For generality, we assume the adversary could launch inference attack with any classifiers. As the precious works~\cite{gong2018attribute}\cite{cai2018collective}, we select different classifiers, including MLP, SVM, k-nearest neighbors (KNN)~\cite{altman1992introduction}, Random forest~\cite{ho1995random}, Adaboost~\cite{friedman2000additive}, Gradient boost~\cite{friedman2000additive}, as attack models to evaluate the robustness of our proposed methods.
We train the attack models using 10\%-90\% users' embeddings and private attribute labels  to infer the other users' sensitive attributes and show the results in Fig.~\ref{fig:DiffAttackYale} and \ref{fig:DiffAttackRochester}. As we can see, for all the six attack models, APGE outperforms the other methods significantly on reducing the Macro-F1 of sensitive attribute prediction for both datasets.
That's because we attempt  to conceal the private information from embeddings with APGE, if the information is removed from embeddings enough, any classifier could not infer the sensitive attribute from the embeddings.


There's another observation, when we only remove the sensitive attribute labels directly from the input attribute matrix (i.e., GAE\_RM), the adversary could predict the other users' privacy with the Macro F1 more than 0.8 on Yale dataset, even if the adversary just obtain  10\% users' sensitive attribute,. That means the users' privacy faces serious threat, if the data owner publishes graph embedding without considering privacy preserving.

\subsection{Effect of the Expansion Layer}

\begin{figure}[t]
\centering
     \subfloat[Impact on utility\label{subfig-1:dummy}]{%
       \includegraphics[width=0.245\textwidth]{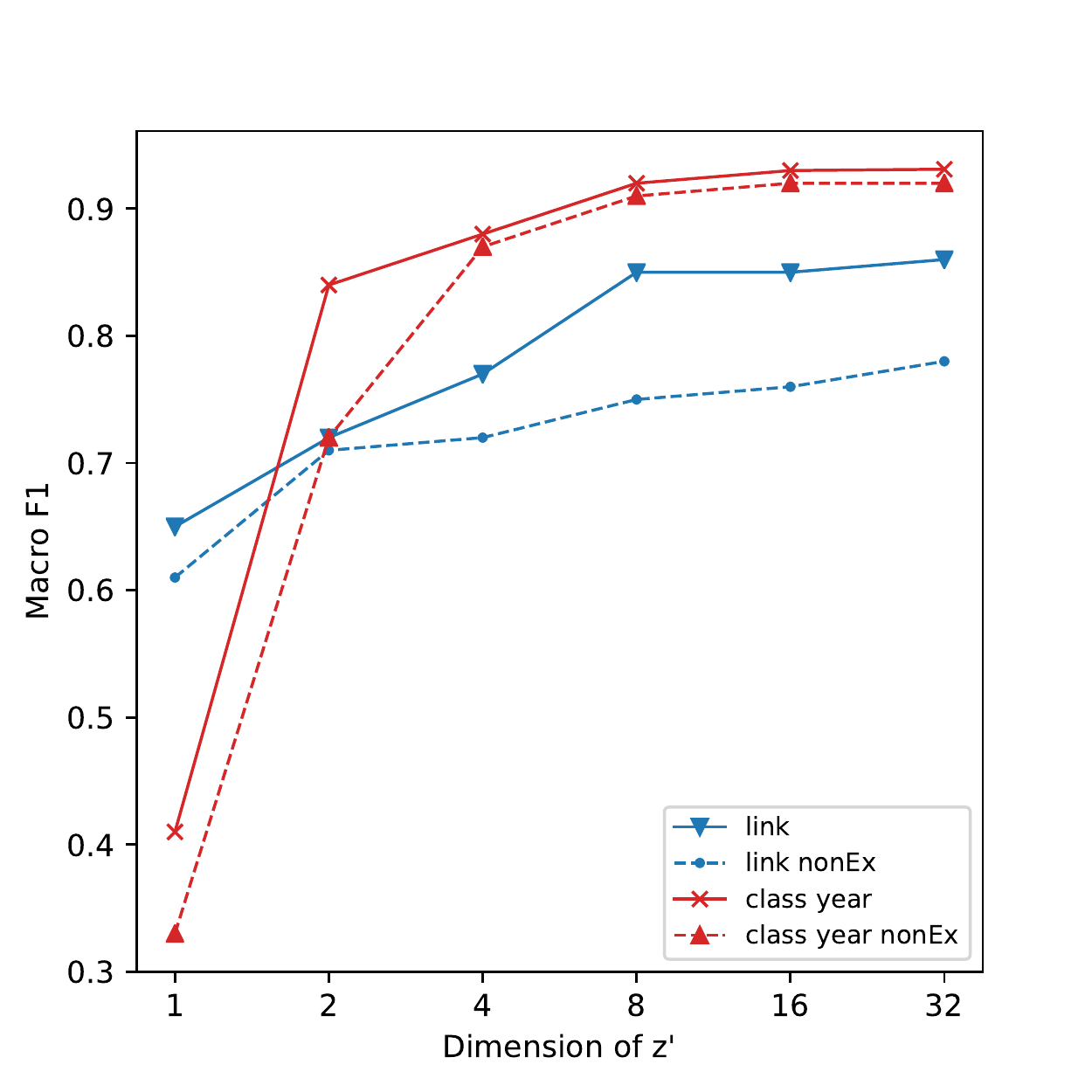}
     }
     \subfloat[Impact on privacy\label{subfig-2:dummy}]{%
       \includegraphics[width=0.245\textwidth]{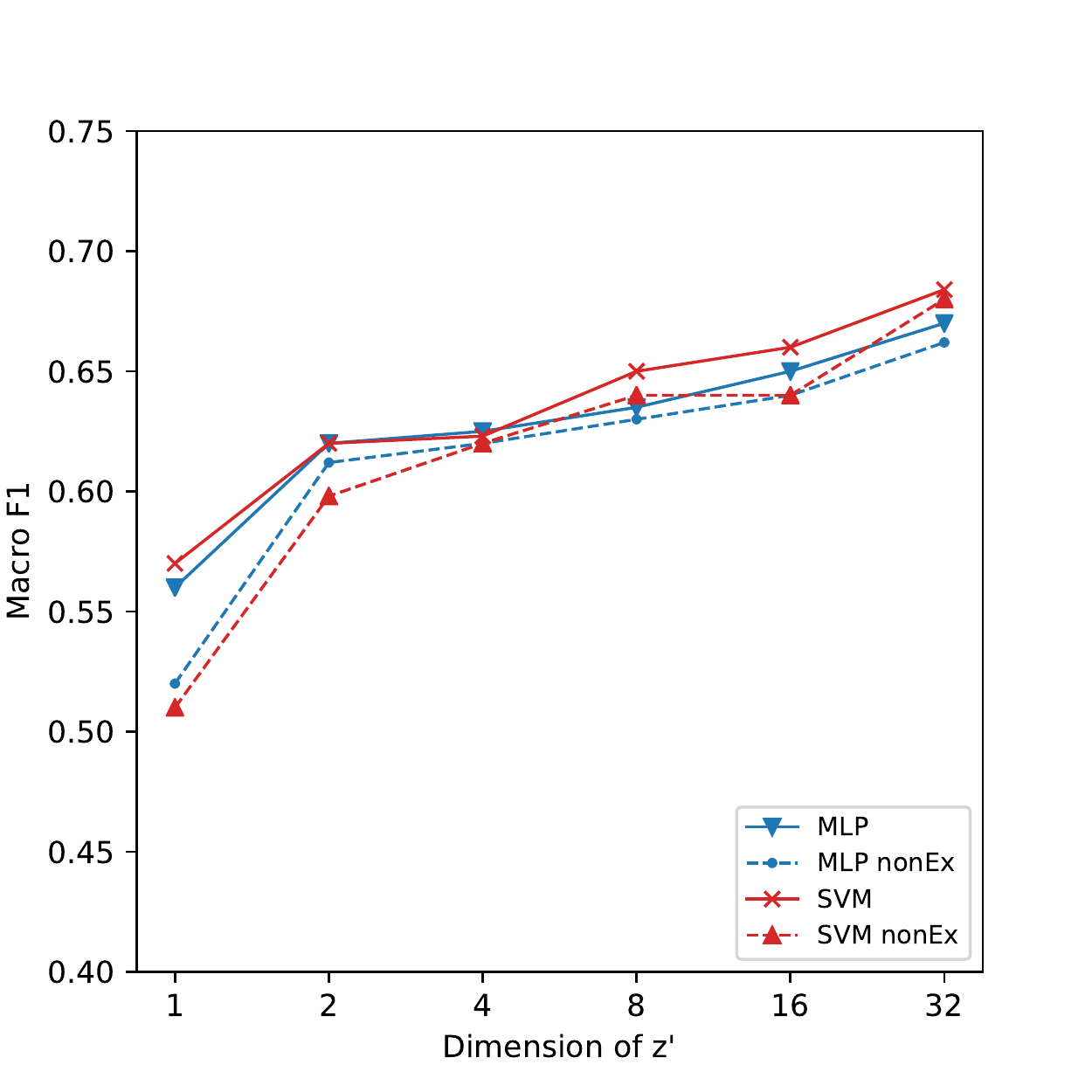}
     }
     \caption{Impact of the Expansion layer}
     \label{fig:ExpanImpact}
 \end{figure}

As introduced in the section \ref{sec:APDGE}, we expand $\mathbf{Z}'$ to $\mathbf{Z}$ via an expansion layer in APDGE to improve the embedding performance.  Since the structure of the Obfuscator of APGE is the same as the ADPGE, the expansion layer is also used in APGE  as shown in Fig.~\ref{Fig:APGE}. In this section, we remove the expansion layer of APGE and directly input  $\mathbf{Z}'$ to Attacker and Decoder to evaluate the impact of the expansion layer. After finishing training the non-expansion layer model (non-Ex model for short), we obtain the  $\mathbf{Z}'$ as the embeddings. We qualify the utility and privacy of the embeddings learned by APGE and  non-Ex model on Rochester, and show the results in Fig.~\ref{fig:ExpanImpact}. We can observe that the  link prediction performance of APGE is significantly larger than the non-Ex model.  The dimension of  $\mathbf{Z}'$  must be small enough to disentangle the private information. Therefore,  When we learn the embeddings via the non-Ex model, the dimension of embeddings (i.e., the dimension of $\mathbf{Z}'$) is too small to sufficiently contain the utility information.
On the other hand, as a lot of private information is concealed in both APGE and the non-Ex model,  even if the embeddings' dimension is small, the embeddings could contain almost all the remaining private information. Therefore, if we remove the expansion layer,  the inference accuracy of sensitive attributes does not reduce remarkably, as shown in Fig. \ref{fig:ExpanImpact}(b).

\section{Related Work}
In this section, we briefly survey the existing graph embedding methods and introduce the state-of-the-art privacy preservation research on graph.
\subsection{Graph Embedding Method}
DeepWalk \cite{perozzi2014deepwalk} is the early work to learn graph representation by generating the node context via random walk and mapping nodes to vectors based on skip-gram \cite{mikolov2013efficient}. Node2vec \cite{grover2016node2vec} exploits a new method to generate node context with considering both local and global structure information and then embeds graph using DeepWalk. The above methods just utilize topology information, and some works have been proposed to improve embedding performance taking into account both the structure and attribute information. TirDNR~\cite{pan2016tri} combines DeepWalk and Doc2Vec~ \cite{dai2015document} to learn the inner-node relationship, node-word correspondence, and word-label correlation. UPP-SNE~\cite{zhang2017user} uses the attribute matrix of social network as input and the node context as label to supervised training model. Different from skip-gram architecture, GAE~\cite{kipf2016variational} is a graph convolutional network via an approximation of spectral graph convolutions for learning the graph representation and outperforms skip-gram architecture. Recently, ARGA \cite{pan2018adversarially} is proposed to add regularization to GAE via an adversarial framework. Unfortunately, these above graph embedding methods do not consider privacy preservation; that is, adversaries can infer the user's sensitive attribute from the embedding results.

\subsection{Privacy Preservation on Graph}
The only existing works to preserve privacy on graph embedding is to achieve the link information differential privacy. That is, even if we add or delete an edge from the original graph, there is no significant statistic difference between the embeddings of the original graph and the changed graph .
Xu {\em et al.}~\cite{xu2018dpne} adopts the objective perturbation mechanism~\cite{chaudhuri2011differentially} on the loss function of the embedding model to achieve differential privacy.  Zhang and Ni ~\cite{zhang2019graph}  propose  a Lipschitz condition~\cite{jha2013testing} on the objective function of the embedding model and a gradient clipping strategy to ensure the differential privacy on link information. However, both of the above approaches do not consider to defend inference attack on users' sensitive attributes.

There has been a few attempts to prevent inference attack with directly sanitizing  graph data. Jia and Gong~\cite{jia2018attriguard} build a set of noise via adversarial machine learning and randomly select the noise to mislead the inference attacker. However, they only manipulate the users' attributes and ignore the  linkages, and thus their method cannot resist the inference attack utilizing topology information well. Cai {\em et al.}~\cite{cai2018collective} first propose an inference attack method with the mixture of non-sensitive attribute and link relationship and then design a privacy-preserving method by removing or perturbing users' attributes and links that are closely related to private attribute. He {\em et al.}~ \cite{he2018latent} obtain the data-sanitization strategy by solving an optimization problem, which can balance the trade off between utility and privacy. However, these prior works do not integrate the processes of privacy protection and graph embedding, and thus is not suitable for privacy-preserving embedding on graphs.

\section{Conclusion}
This paper proposes a novel graph embedding framework that produces node representations with users' privacy preserved. To the best of our knowledge, this is the first graph embedding method that integrates privacy protection into GCN. We introduce two different mechanisms for privacy protection in graph embedding: latent factor disentangling and adversarial training. The resulting algorithm APGE which integrates the two mechanisms into one end-to-end pipeline demonstrates superior performance as compared to the state-of-the-arts in privacy protection while retaining similar accuracy in link prediction and users' utility classification.

\bibliographystyle{IEEEtran}
\bibliography{bibfile}







\begin{IEEEbiography} [{\includegraphics[width=1in,height=1.25in,clip,keepaspectratio]{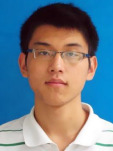}}]{Kaiyang Li}
received the M.S. degree from the School of Energy Science and Engineering University of Electronic Science and Technology of China, Chengdu, China, in 2016. He is currently a Ph.D. student in the School of Computer Science and Engineering, University of Electronic Science and Technology of China. His research interests include privacy and machine learning.
\end{IEEEbiography}

\begin{IEEEbiography}[{\includegraphics[width=1in,height=1.25in,clip,keepaspectratio]{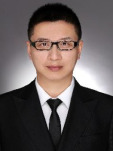}}]{Guangchun Luo}
received the Ph.D. degree in computer science from University of Electronic Science and Technology of China, Chengdu, China, in 2004. He is currently a professor of computer science at the  University of Electronic Science and Technology of China. He has published over sixty journal and conference papers in his fields. His research interests include computer networks and big data.
\end{IEEEbiography}

\begin{IEEEbiography}[{\includegraphics[width=1in,height=1.25in,clip,keepaspectratio]{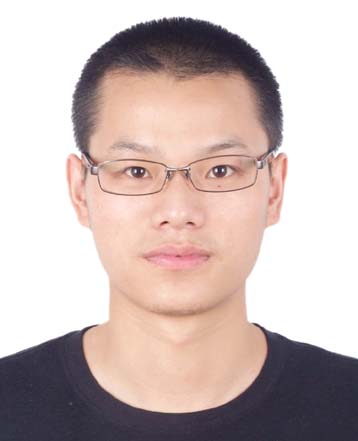}}]{Yang Ye}
received the M.S. degree in computer science  from Huazhong University of Science and Technology, Wuhan, Hubei, China, in 2014. He is currently working toward the Ph.D. degree in the Department of Computer Science, Georgia State University, Atlanta, GA, USA. His research interests include deep learning and data minig.
\end{IEEEbiography}

\begin{IEEEbiography}[{\includegraphics[width=1in,height=1.25in,clip,keepaspectratio]{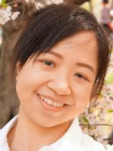}}]{Wei Li}
currently is an Assistant Professor in the Department
of Computer Science at Georgia State University.
Dr. Li received her Ph.D. degree in Computer
Science from The George Washington University in
2016 and her M.S. degree in Computer Science from
Beijing University of Posts and Telecommunications
in 2011. She won the Best Paper Awards in ACM
MobiCom CRAB 2013 and WASA 2011, respectively.
Her current research mainly spans the areas
of security and privacy for the Internet of Things and
Cyber-Physical Systems, secure and privacy-aware
computing, Big Data, Blockchain, Game Theory, and algorithm design and
analysis. She is a member of IEEE and ACM.
\end{IEEEbiography}

\begin{IEEEbiography}[{\includegraphics[width=1in,height=1.25in,clip,keepaspectratio]{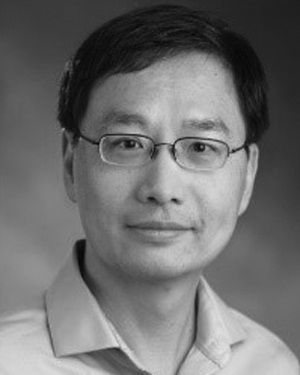}}]{Shihao Ji}
 received the PhD degree in electrical and computer engineering from Duke University, in 2006. After that, he was an research associate with Duke for about 1.5 years. He is an associate professor with the Computer Science Department, Georgia State University. His principal research interests lie in the area of machine learning and deep learning with an emphasis on high-performance computing. He is interested in developing efficient algorithms that can learn from a variety of data sources (e.g., image, audio, and text) on a large scale and automate decision-making processes in dynamic environments. Prior to joining GSU, he spent about 10 years in industry research labs.
\end{IEEEbiography}

\begin{IEEEbiography}[{\includegraphics[width=1in,height=1.25in,clip,keepaspectratio]{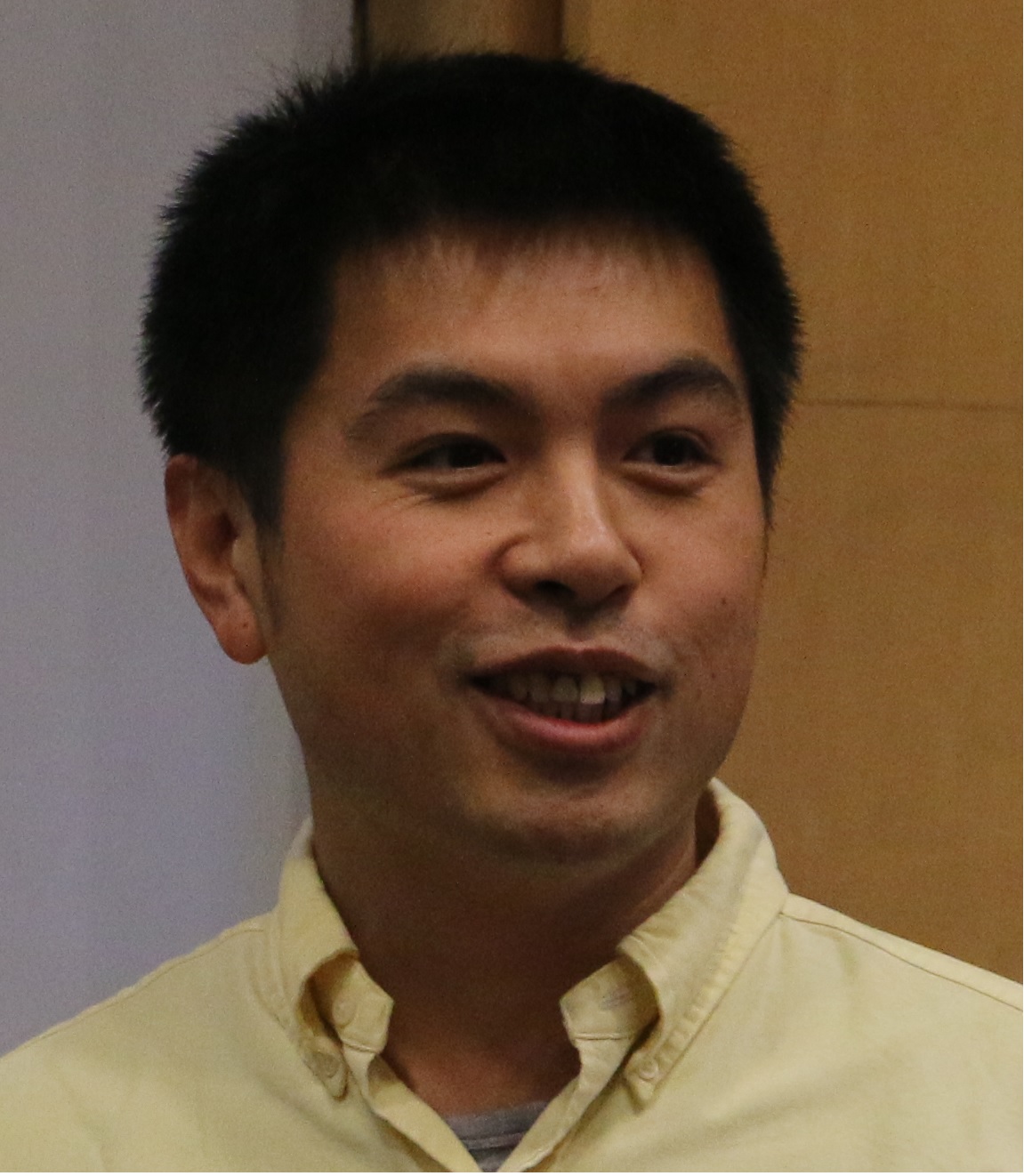}}]{Zhipeng Cai}
received the B.S. degree from the
Beijing Institute of Technology, Beijing, China, and
the M.S. and Ph.D. degrees in computing science
from the University of Alberta, Edmonton, AB,
Canada. He is currently an Associate Professor
with the Department of Computer Science, Georgia
State University. His research interests include
networking, privacy, and big data. He has received
an NSF CAREER Award. He is an editor/guest
editor of the IEEE Transactions on Knowledge and
Data Engineering, IEEE Transactions on Vehicular
Technology, IEEE Networking Letters, Algorithmica, Theoretical Computer
Science, the Journal of Combinatorial Optimization, and the IEEE/ACM
Transactions on Computational Biology and Bioinformatics. He is a senior
member of the IEEE.
\end{IEEEbiography}

\end{document}